\documentclass[10pt,twocolumn,letterpaper]{article}

\usepackage[pagenumbers]{cvpr} 

\usepackage{graphicx}
\usepackage{amsmath}
\usepackage{amssymb}
\usepackage{booktabs}
\usepackage{algorithm}
\usepackage{algpseudocode}
\usepackage{multirow}
\usepackage[dvipsnames]{xcolor,colortbl}
\usepackage{chapterbib}
\usepackage{epstopdf}
\usepackage{stackengine}

\usepackage[accsupp]{axessibility}  

\newenvironment{packedEnum}{
\begin{enumerate}
  \setlength{\itemsep}{1pt}
  \setlength{\parskip}{0pt}
  \setlength{\parsep}{0pt}
}{\end{enumerate}}

\DeclareMathAlphabet{\mbf}{OT1}{ptm}{b}{n}

\newcommand{\norm}[1]{\left\Vert#1\right\Vert} 

\newcommand{\mrm}[1]{\mathrm{#1}}
\newcommand{\trans}{{\ensuremath{\mathsf{T}}}}

\definecolor{cvprblue}{rgb}{0.21,0.49,0.74}
\usepackage[pagebackref,breaklinks,colorlinks,allcolors=cvprblue]{hyperref}
\usepackage[capitalize]{cleveref}
\crefname{section}{Sec.}{Secs.}
\Crefname{section}{Section}{Sections}
\Crefname{table}{Table}{Tables}
\crefname{table}{Tab.}{Tabs.}

\def\paperID{8506} 
\def\confName{CVPR}
\def\confYear{2025}

\title{Large Self-Supervised Models Bridge the Gap in Domain Adaptive \\ Object Detection}

\author{Marc-Antoine Lavoie \qquad Anas Mahmoud \qquad Steven L. Waslander \\
University of Toronto Robotics Institute \\
{\tt\small \{marc-antoine.lavoie, anas.mahmoud, steven.waslander\}@robotics.utias.utoronto.ca}}

\begin{document}
\maketitle
\begin{abstract}
The current state-of-the-art methods in domain adaptive object detection (DAOD) use Mean Teacher self-labelling, where a teacher model, directly derived as an exponential moving average of the student model, is used to generate labels on the target domain which are then used to improve both models in a positive loop. This couples learning and generating labels on the target domain, and other recent works also leverage the generated labels to add additional domain alignment losses. We believe this coupling is brittle and excessively constrained: there is no guarantee that a student trained only on source data can generate accurate target domain labels and initiate the positive feedback loop, and much better target domain labels can likely be generated by using a large pretrained network that has been exposed to much more data. Vision foundational models are exactly such models, and they have shown impressive task generalization capabilities even when frozen. We want to leverage these models for DAOD and introduce DINO Teacher, which consists of two components. First, we train a new labeller on source data only using a large frozen DINOv2 backbone and show it generates more accurate labels than Mean Teacher. Next, we align the student's source and target image patch features with those from a DINO encoder, driving source and target representations closer to the generalizable DINO representation. We obtain state-of-the-art performance on multiple DAOD datasets. Code available at \url{https://github.com/TRAILab/DINO_Teacher}.
\end{abstract}

\section{Introduction}
\label{sec:introduction}

Modern deep learning methods can achieve strong performance on a wide range of benchmark datasets for many computer vision tasks, including classification, object detection, and segmentation. However, this performance often degrades significantly when tested on data outside the training distribution. Domain adaptation addresses this issue by transferring the performance of a network trained on a labelled source dataset to an unlabelled target dataset \cite{ganin2015unsupervised}. In this work, we specifically consider the domain adaptive object detection (DAOD) task.

Two common techniques for achieving domain adaptation are domain invariance and self-labelling. Domain invariance attempts to minimize the difference between features from the source and target domains, either by using an adversarial strategy with a domain discriminator \cite{ganin2015unsupervised, hoffman2016fcns, tzeng2017adversarial, chen2018domain, saito2019strong, vs2021mega} or by minimizing some distance metric between domains \cite{courty2014optimal,kang2019contrastive,li2022scan}.
\begin{figure}[t]
  \centering
    \includegraphics[width=0.46\textwidth]
    {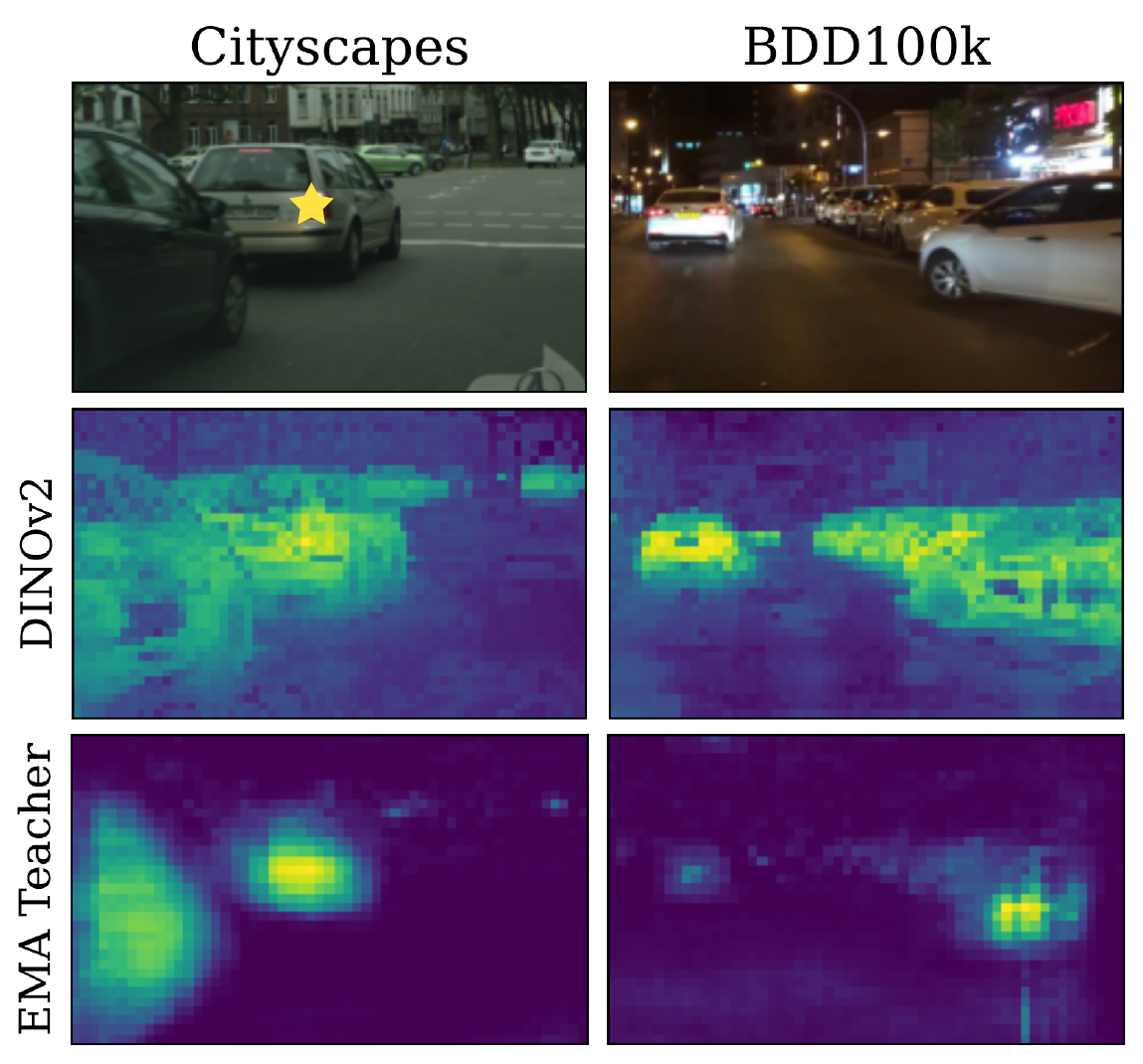}
    	\caption{\textbf{Cosine similarity of patch feature from Cityscapes to BDD100k.}. We evaluate the similarity between the yellow star region in the Cityscapes image and all other regions in both images. Compared to the EMA Teacher, DINOv2 generates semantically stable features across domains, justifying the choice to use it for feature alignment and as a frozen backbone for our labeller.}
    	\label{fig:feat_align}
\end{figure}
Self-labelling methods instead attempt to generate their own pseudo-labels on the target dataset. The most common framework for this is Mean Teacher \cite{tarvainen2017mean,liu2021unbiased,deng2021unbiased}, where the labeller is directly derived from the model being trained. This is a bootstrapping method, where the model selects confident proposals to use as ground truth to train on the target domain, thereby improving its performance and generating more and higher quality pseudo-labels at the next iteration. Because the performance of this self-labelling approach is sensitive to the choice of pseudo-labels to train on, many methods attempt to refine label selection by going beyond simple class confidence thresholding \cite{li2022rethinking,kar2023revisiting,deng2023harmonious}.

At the intersection of the two sets of methods are works that use self-generated labels to refine domain invariance approaches \cite{li2022scan,zhao2022task,cao2023contrastive}. By using labels, it is possible to perform alignment at the proposal \cite{cao2023contrastive} or pixel levels \cite{xie2023sepico} and constrained to instances of the same class, improving on the earlier image-level methods. However, these methods are limited by the performance of the teacher network and can perform poorly if some classes are undersampled in the target domain. These issues make it particularly difficult to adapt to small datasets with significant domain gaps, such as severe adverse weather in autonomous driving.

Recently, large-scale vision foundation models (VFMs), whether purely vision-based (DINOv2 \cite{oquab2024dinov2}, SAM \cite{kirillov2023segment}) or combining vision and language (CLIP \cite{radford2021learning}), have become widely available. These models combine large architectures with self-supervised pre-training on large-scale training datasets and have shown impressive performance on multiple computer vision tasks, including as guides in augmentation and generation methods for domain adaptation \cite{liu2024unbiased} and generalization \cite{fahes2023poda,yang2020fda,benigmim2024collaborating}. Given this, we believe the current Mean Teacher paradigm for DAOD is outdated. While there are obvious applications of domain adaptation on small architectures for robotics or autonomous driving, and pseudo-labelling remains a powerful strategy, we believe it is no longer reasonable to simply use a duplicate of the student model as the teacher. By decoupling labelling from training the student model, we can use more modern methods to solve domain adaptation while maintaining the size of the trained model. Here, we propose to use vision foundation models with minimal adjustments as the source of the labels, highlighting the usefulness of VFMs in multiple adaptation tasks. Similarly, while class-conditional alignment between domains can be beneficial, we choose to decouple cross-domain alignment from accurate instance labels. To this end, we propose to align student representations with those generated from a VFM backbone. Features from these models have been shown to generalize well to different tasks and new target domains. We illustrate these consistent features for similar objects across different datasets in \cref{fig:feat_align}. 


 Thus, we propose DINO Teacher (DT) for the DAOD task which exploits large vision foundation models in two ways. First, we show that a simple detector added to the frozen encoder of a large-scale model and trained only on source data is a better target domain labeller than Mean Teacher, demonstrating the drawbacks of the symmetric architecture and limited training data of the Mean Teacher approach. Next, we propose to use the feature space of a large-scale VFM as a proxy for domain alignment. By aligning the student backbone features with those generated by a VFM independently for source and target images, we drive student source and target features to be closer. Our key contributions are: 
\begin{itemize}
    \item We show that using pseudo-labels from a simple detector trained on a frozen large self-supervised encoder gives substantially more accurate pseudo-labels than a student-derived teacher, and using these new labels greatly improves performance.
    \item Next, we show that aligning the patch-level output of the small student backbone with the frozen large VFM on both source and target images drives the model to learn features that generalize better without using any labels, decoupling feature alignment from pseudo-label quality.
    \item We achieve state-of-art performance on multiple DAOD benchmarks, with improvements of $+7.6\%$ on BDD100k and $+2.3\%$ on Foggy Cityscapes. We also present new results on the ACDC test splits, which cover more extensive adverse weather conditions, and significantly improve our baseline.
\end{itemize}






\section{Related Works}
\label{sec:rel_works}

\subsection{Domain Adaptive Object Detection}
Domain adaptive object detection (DAOD) builds on earlier works in domain adaptation for image classification. The DAOD task is to adapt a detector trained on a labelled source domain to an unlabelled target domain. A common approach for domain adaptation is self-labelling or pseudo-labelling. These methods generate labels for the unsupervised target domain by sampling confident proposals from a model initially trained on the source domain only. This often uses the Mean Teacher framework (MT) \cite{liu2021unbiased,deng2021unbiased}, where an iteration-averaged version of the model generates more robust labels. This framework forms the baseline of many recent works \cite{deng2023harmonious,cao2023contrastive,kennerley2024cat,li2024react}. Because this self-labelling approach attempts to bootstrap itself into generating more and better labels, it is sensitive to errors or biases in the generated pseudo-labels. Many works try to reduce the bias against rare classes by changing the selection threshold \cite{li2022rethinking,nie2023adapting,kar2023revisiting} or oversampling instances of rare classes \cite{zhang2022semi,kennerley2024cat}. Other methods improve box localization by estimating box uncertainty and using it to select or reweigh boxes \cite{xu2021end,chen2022learning,deng2023harmonious}. We believe that the coupling between labelling and the performance of the student model in the Mean Teacher framework limits the quality of generated labels. Instead, we propose to decouple label generation from solving the domain adaptation task on the student model by training an independent labeller to provide the best possible pseudo-labels to the student. Because we still need to generate the best labels to train the student, some of the recent improvements such as oversampling rare classes \cite{kennerley2024cat} could be complimentary to our framework. 

Another common strategy in domain adaption is domain invariance. This was first formulated as an adversarial loss \cite{ganin2015unsupervised,tzeng2017adversarial} that pushes image-level backbone features to contain no domain information. Other methods obtain domain invariant features by instead directly minimizing a distance measure between the feature representations of the source and target domains. Some of the distance measures used include contrastive loss \cite{kang2019contrastive,xie2023sepico,cao2023contrastive}, graph alignment \cite{li2022scan,liu2022towards,li2022sigma,gao2023csda} or optimal transport \cite{courty2014optimal}. The simplest invariance or alignment methods operate on image-level features, but there is no guarantee that image-level alignment leads to the instance-level alignment needed for detection and segmentation tasks \cite{kumar2018co}. Alternatively, alignment methods can be combined with the self-labelling approaches discussed above, allowing instance-level and class-conditional alignment \cite{chen2018domain,xu2020cross,li2022sigma,zhao2022task,cao2023contrastive,zhou2023unsupervised}. However, coupling the alignment term with self-labelling leads to a bind: good labels are required for robust alignment, but domain alignment is also required to generate good labels. Instead, we propose to indirectly align the domains by encouraging the feature space of the student model to be similar to the feature space of a self-supervised VFM on both the source and the target images without requiring any labels. 

Other methods use stronger augmentations, reducing overfitting on the source domain and the limited generated target domain labels. This includes implementations of Mixup \cite{zhang2018mixup,zhang2022semi,kennerley20232pcnet} and CutMix \cite{yun2019cutmix,mattolin2023confmix}, the weak-strong approach of Unbiased Teacher \cite{liu2021unbiased} or various frequency-space transforms \cite{yang2020fda,xu2023multi,liu2024unbiased}, which assume that certain frequency bands or that the magnitude spectrum are domain-specific and can be perturbed or transferred across images. Similarly, some methods use generative models to learn a mapping between source and target domain \cite{zhu2017unpaired,shan2019pixel,deng2021unbiased} to transform labelled source images into target-like images. A final set of methods uses masked reconstruction \cite{zhao2023masked,hoyer2023mic} in the feature space to learn better target domain features. We believe these methods could be complementary to ours.

\subsection{Large Models in Domain Adaptation and Generalization}
Recently, many methods have begun leveraging VFMs for domain adaptation. Some methods attempt to refine large pre-trained models by fine-tuning only a few parameters \cite{hulora,chen2022adaptformer} or by modifying the input prompt to the encoder \cite{du2024domain,ge2023domain,singha2023ad,wei2024stronger}. Other methods are similar to the generative approaches discussed above, with methods such as CLIPStyler \cite{kwon2022clipstyler}, PØDA \cite{fahes2023poda}, FAMix \cite{fahes2024simple} and ULDA \cite{yang2024unified} all showing that given aligned vision-language embeddings, it is possible to synthesize new target-like images or image features by replicating a domain shift encoded by text in the aligned image feature space, allowing adaptation without requiring any target images. While all these methods are interesting avenues of research, we focus on the standard DAOD task that attempts to maximize performance on an unlabelled target dataset given a relatively small model architecture (\eg VGG16 \cite{simonyan2014very}), which are essential when deploying networks in the real world.

Closer to our work, UFR \cite{liu2024unbiased} recently used Segment Anything \cite{kirillov2023segment} to refine instance masks for augmentation but still trains and evaluates on smaller architectures. While augmentation methods are useful for domain adaptation, self-labelling methods generally show stronger adaptation performance. Given this, we investigate the use of VFMs as labellers. As mentioned previously, one of the advantages of these large models is their generalizability to other downstream tasks. A recent work, REIN \cite{wei2024stronger}, shows that fully frozen backbones are nearly on par with fully fine-tuned ones for domain generalized semantic segmentation, confirming strong feature generalizability for computer vision tasks. We want to leverage this capability to use a large model as a labeller on the unlabelled target domain data and as an alignment target for the student network.

\section{Method}
\label{sec:method}
\begin{figure*}[t]
  \centering
    \includegraphics[width=0.78\textwidth]
    {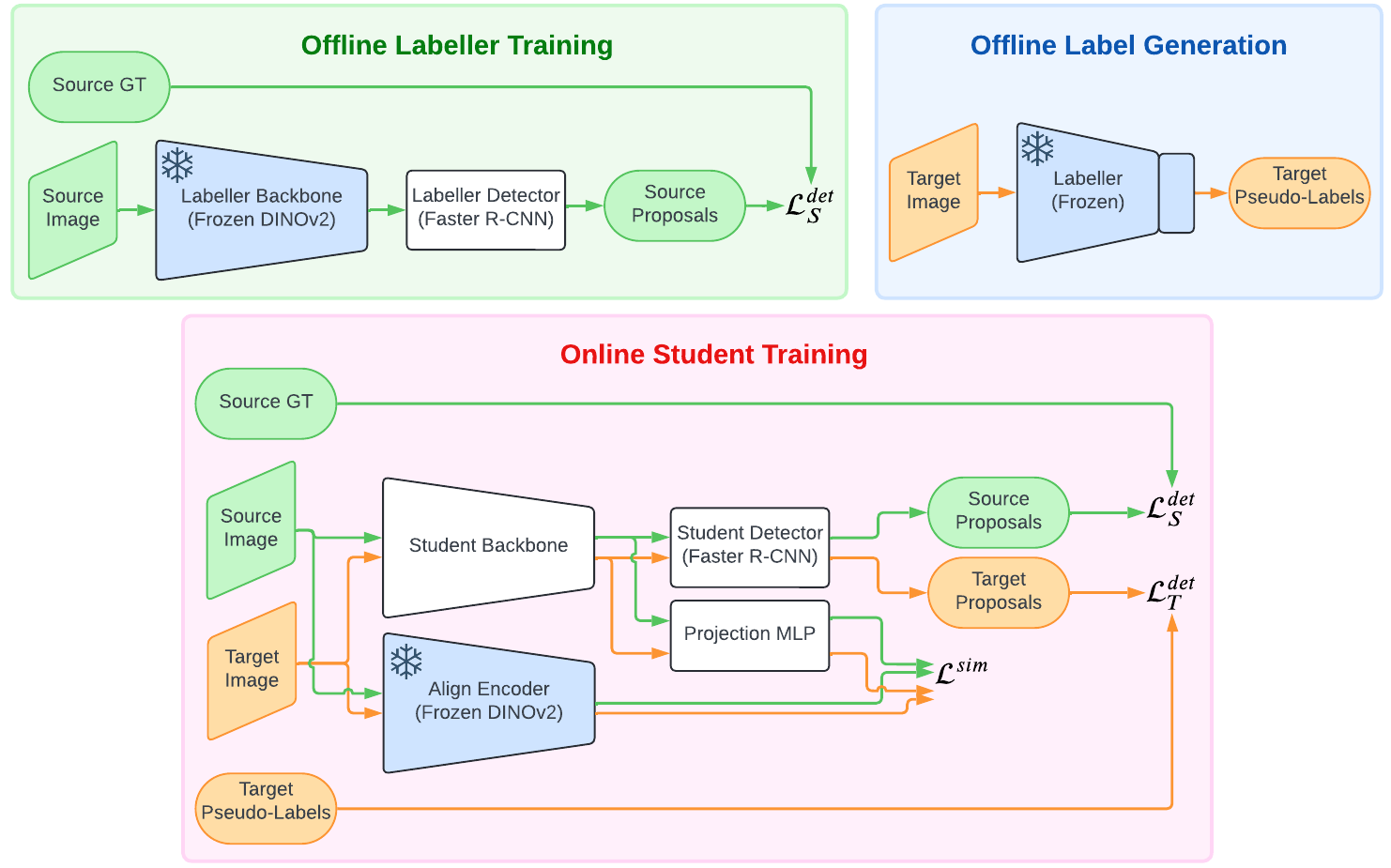}
    	\caption{\textbf{Diagram of our proposed method.} {\color{ForestGreen}\textbf{Offline Labeller Training:}} we add a detector head to a frozen DINOv2 encoder and train it with source images only. {\color{Blue}\textbf{Offline Label Generation:}} we combine and freeze the labeller backbone and detector, and generate target pseudo-labels. {\color{Red}\textbf{Online Student Training:}} We train a student network using source ground truth boxes and target pseudo-labels, and align patch features to a frozen DINOv2 encoder. During inference, the alignment encoder and projection MLP are not used.}
    	\label{fig:schematic}
\end{figure*}

In this section, we summarize the domain adaptive object detection problem, present the baseline method and explain our contributions. We build primarily on the Adaptive Teacher \cite{li2022cross} framework and frequently contrast our design choices with that method. \cref{fig:schematic} summarizes our proposed DINO Teacher framework.

\subsection{Problem Definition}
\label{sec:method_prob_def}
The problem we consider is domain adaptive object detection, where we want to transfer performance from a labelled source domain, $\mathcal{D}_{S}$, to an unlabelled target domain, $\mathcal{D}_{T}$, which shares the same classes. We thus have images, $X_{S}$, bounding boxes, $B_{S}$, and class labels, $Y_{S}$, on the source domain while we only have images, $X_{T}$, on the target domain. To train on target images, we first have to generate target pseudo-labels, $\tilde{B}_{T}$ and $\tilde{Y}_{T}$. We use Faster R-CNN \cite{ren2015faster} as our detector, and the training loss is
\begin{align}
    \mathcal{L}^{det} &= \mathcal{L}^{det}_{S}\left(X_{S},B_{S},Y_{S}\right) +\mathcal{L}_{T}^{det}\left(X_{T},\tilde{B}_{T},\tilde{Y}_{T}\right) \label{eq:mt_loss}  \\
    &= \mathcal{L}_{S}^{rpn}\left(X_{S},B_{S},Y_{S}\right) + \mathcal{L}_{S}^{roi}\left(X_{S},B_{S},Y_{S}\right) \nonumber \\
    &+ \mathcal{L}_{T}^{rpn}\left(X_{T},\tilde{B}_{T},\tilde{Y}_{T}\right) + \mathcal{L}_{T}^{roi}\left(X_{T},\tilde{B}_{T},\tilde{Y}_{T}\right),\nonumber
\end{align}
where $\mathcal{L}^{rpn}$ and $\mathcal{L}^{roi}$ are the standard Faster R-CNN region proposal network (RPN) box proposal and Region of Interest (RoI) box refinement and classification losses.

\subsection{Mean Teacher Framework}
\label{sec:method_mean_teacher}
In the Mean Teacher framework \cite{deng2021unbiased}, a student model, $f$, trained on the source data is used to instantiate a teacher, $\bar{f}$, to generate pseudo-labels on the target domain. This teacher is a duplicate of the original network whose parameters, $\bar{\theta}$, are updated as an exponential moving average (EMA) of the student parameters, $\theta$. The teacher is thus essentially an ensemble over iterations of the student training process
\begin{align}
    \bar{\theta} &\leftarrow \alpha \bar{\theta} + \left(1 - \alpha \right) \theta, \label{eq:ema}
\end{align}
where $\alpha$ is the EMA decay factor. To generate pseudo-labels, each target image is augmented using two different augmentations. The first augmented image is fed to the teacher network, which generates pseudo-labels by thresholding box proposals with class probabilities above a given threshold, $\delta$. The second augmented image is sent to the student, and the thresholded teacher proposals are used as ground truth for the student. Pseudo-label training on the target domain starts after some initialization phase of $n^{initPL}$ iterations training exclusively on source images. Unbiased Teacher also proposes to improve learning by using weak-strong augmentations \cite{liu2021unbiased}. Simple augmentations (crop, flip) are applied to the teacher images to make labelling easy, while the student images have stronger augmentations (blurring, colour jitter, grayscaling, Cutout \cite{devries2017improved}).

\subsection{Foundation Models for Pseudo-Labelling}
\label{sec:method_large_model_labels}
Our method generally follows the same self-labelling strategy, but instead of using the time-averaged student, $\bar{f}$, as a teacher, we propose to use another model entirely, $f^{\mrm{big}}$. A key issue with using the student model as a starting point for the teacher is that the model operates outside its training data distribution when labelling target data. Because DINOv2 is trained using self-supervised losses on a large corpus of data \cite{oquab2024dinov2}, we hypothesize that its feature representation will generalize well, including over the considered domain gap from source to target. Thus, we propose to use DINOv2 as the encoder for a new labeller, and we train an object detection head on top of the frozen encoder using only the source data. We confirm that this leads to a transferable detector by evaluating the performance of the DINO labeller on the unseen target data in \cref{tab:student_vs_teacher}.

We use DINOv2 ViT-G \cite{dosovitskiy2020image} as our frozen encoder backbone and use a Faster R-CNN detection head. We do not require that the detection head be identical to the student network's because we only use the output box proposals (see \cref{fig:schematic}) when training the student, but we use the same detector as the student model to highlight the impact of changing the backbone. We train the labeller detector with the Faster R-CNN loss $\mathcal{L}^{det}_{S}$ presented in \eqref{eq:mt_loss}. 

We generate all pseudo-labels $\tilde{B}_{T}$ and $\tilde{Y}_{T}$ for the target domain training examples as a preprocessing step in a single forward pass over the target training data. To select pseudo-labels, we threshold the bounding box proposals according to class probability similar to Mean Teacher. In our architecture, $f^{\mrm{big}}$ replaces the student-derived teacher $\bar{f}$ for labelling, but we still generate and evaluate performance on the EMA model $\bar{f}$ as it is significantly better than the student $f$ on the detection task. Model ensembling \cite{laine2017temporal} improves performance outside of its use as a teacher and is a common training strategy in recent vision models \cite{caron2021emerging,wang2023yolov7}.

\subsection{Foundation Models for Feature Alignment}
\label{sec:method_large_model_alignment}

As discussed in \cref{sec:rel_works}, existing domain invariance methods operate in two ways. First, they can align at the image level, which does not guarantee alignment at the instance level and can be dominated by background information, but which does not require labels. Alternatively, they can use pseudo-labels to match semantically related regions or instances for alignment, which generally improves performance \cite{zhao2022task,li2022sigma} at the cost of relying on accurate generated labels. We propose an alternative alignment method that aligns semantically similar regions without using generated labels by using a proxy alignment target, to which we can independently align the features of the student model from both source and target images and without having to match regions across domains.

\begin{figure*}[ht]
    \centering
    \subcaptionbox{w/o DINO alignment \label{fig:subtsne_wo_align}}[0.29\textwidth]{\includegraphics[width=0.29\textwidth]{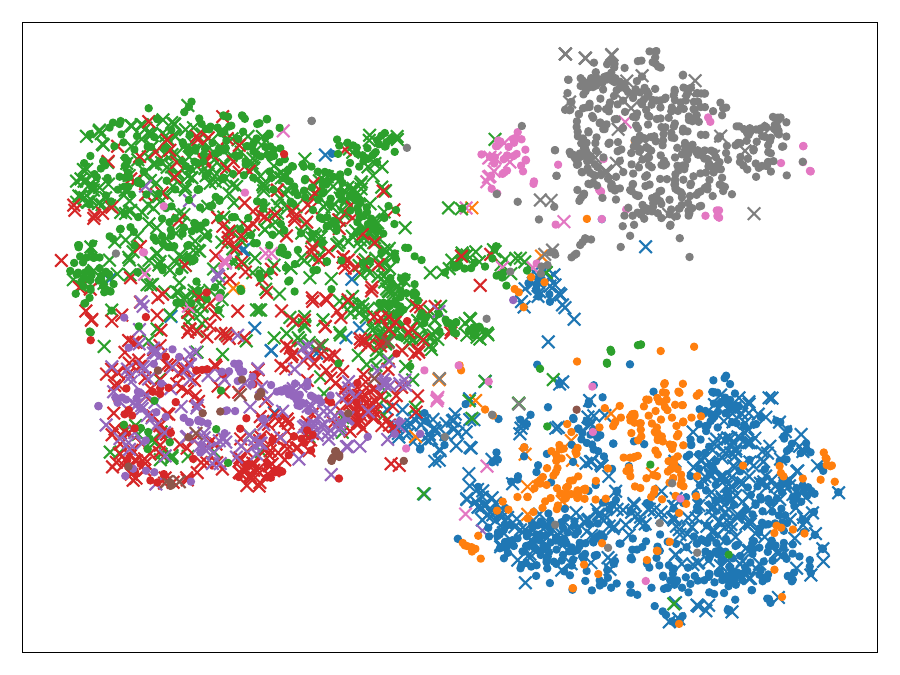}}
    \subcaptionbox{w/ DINO alignment  \label{fig:subtsne_w_align}}[0.29\textwidth]{\includegraphics[width=0.29\textwidth]{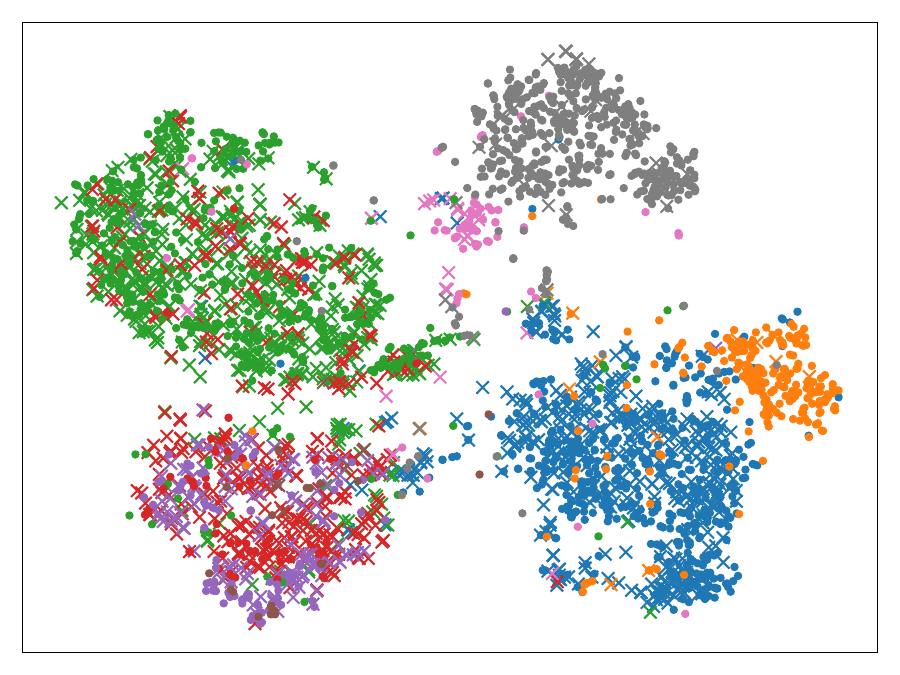}}
    \subcaptionbox{DINOv2 \label{fig:subtsne_dino}}[0.29\textwidth]{\includegraphics[width=0.29\textwidth]{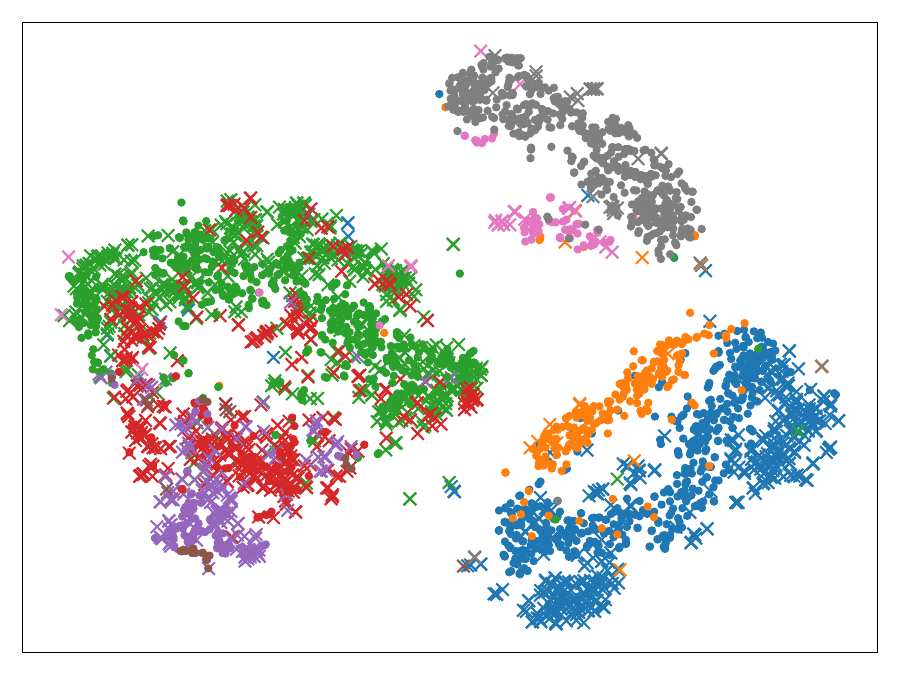}}
    \includegraphics[width=0.093\textwidth]{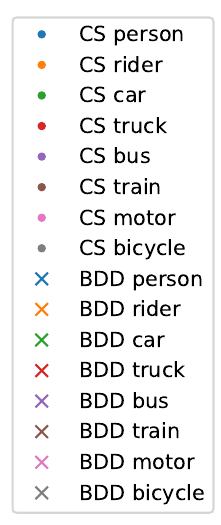}
\caption{\textbf{t-SNE of the backbone instance-level embeddings across domains.} Subfigures \subref{fig:subtsne_wo_align} and \subref{fig:subtsne_w_align} are taken from VGG16 features after 20k training iterations of supervised training on source only (Cityscapes), following the protocol defined in \cref{sec:experiments}. Without alignment \subref{fig:subtsne_wo_align}, there is confusion between all similar classes. With alignment \subref{fig:subtsne_w_align}, overlap is reduced, particularly between {\color{NavyBlue}\textbf{persons}} ({\color{NavyBlue}{$\bullet$}},{\color{NavyBlue}{$\times$}})  and {\color{Orange}\textbf{riders}} ({\color{Orange}{$\bullet$}},{\color{Orange}{$\times$}}). Pretrained DINOv2 \subref{fig:subtsne_dino} has well separated clusters.}
\label{fig:tsnes}
\end{figure*}

To this end, we use the feature representation of a VFM as our alignment target. Again, we hypothesize that the DINOv2 features will generalize better to the target domain and that the representation of any particular object class will be better defined across domains than the representation of a small student model trained on source data only. To improve the student, we propose to align patch-level features of the student backbone with those of a frozen DINOv2 encoder on individual images of both the source and target domains. 
This drives the student representation of both domains towards the large model's representation, reducing the domain gap. We align patch features of the same image processed by the student and by the DINO encoder, ensuring that we respect local semantic information: the same car patch is encoded by both the student backbone and the DINO alignment teacher. We compare the clustering of the instance-level features of the student network with and without alignment to DINOv2 and the original DINOv2 clustering in \cref{fig:tsnes} using t-SNE \cite{van2008visualizing}. This shows that DINO features have better rare class separation and that aligning the student to DINOv2 improves separation on both the source domain Cityscapes \cite{cordts2016cityscapes} and the target domain BDD \cite{yu2020bdd100k}. 


We take DINOv2 ViT-B \cite{dosovitskiy2020image} as our frozen encoder backbone. Given patch-level DINO teacher features $\mbf{x}^{\mrm{big}}$ and student features $\mbf{x}$, we pass the student features through a 2-layer projection MLP $g(\cdot)$ to match the channel dimensions $C$ and then use bilinear interpolation to match the dimensions $H$ and $W$ of the DINO features. The similarity between the normalized features of both backbones is maximized with the alignment loss
\begin{align}
    \mathcal{L}^{sim} &= \dfrac{1}{NHW} \sum_{k}^{N} \sum_{ij}^{HW} 1 - \dfrac{{\mrm{interp}\left(g\left(\mbf{x}\right)\right)}^{\trans}\mbf{x}^{\mrm{big}}}{\norm{\mrm{interp}\left(g\left(\mbf{x}\right)\right)}_2\norm{\mbf{x}^{\mrm{big}}}_2}, \label{eq:loss_align}
\end{align}
where $N$, $H$ and $W$ are the batch size, height and width of the final DINO encoder feature map, and where we have suppressed the features indices $i,j,k$ for ease of reading. 

In implementation, we align the student and DINO features on the source domain from the start, pushing the initial student representation to be predictive of the DINO representation and to learn the projection MLP. After $n^{initSim}$, we align with both source and target images. Notably, we start aligning on the target domain before using any pseudo-labels, different from Adaptive Teacher which starts both simultaneously. We begin domain alignment early to reduce issues of unaligned gradients from supervised training in different domains.

In contrast to the labeller described in \cref{sec:method_large_model_labels} above, we use the smaller DINOv2 ViT-B as the alignment encoder. While the labeller only runs once on each image of the target training dataset and this is done before training the student, the alignment DINO encoder is used during online training, and we found that inference on the larger ViT models was excessively slow. Our method does not require that the two DINO encoders be the same, and the encoders are never used simultaneously (see \cref{fig:schematic}).

\subsection{Complete Method}
Our full loss combines the pseudo-labelling detection loss \eqref{eq:mt_loss} with the similarity loss \eqref{eq:loss_align}
\begin{align}
    \mathcal{L} &= \mathcal{L}^{det}_{S} + \lambda^{unsup} \mathcal{L}^{det}_{T} + \lambda^{sim} \mathcal{L}^{sim}, \label{eq:full_loss}
\end{align}
where $\lambda^{unsup}$ and $\lambda^{sim}$ are hyperparameters that tune the respective weights of the losses associated with the pseudo-labels and the similarity term.

\section{Experiments}
\label{sec:experiments}

\subsection{Datasets}\label{sec:datasets}
\paragraph{Cityscapes.} All of the reported runs use Cityscapes \cite{cordts2016cityscapes} as the source dataset. Cityscapes is an autonomous driving dataset with 2975 training images and 500 validation images, with detection labels for 8 classes.

\paragraph{Foggy Cityscapes.}
Foggy Cityscapes \cite{sakaridis2018semantic} is a synthesized dataset that simulates foggy weather at three intensities (0.005, 0.01, 0.02) using the original clear images and a depth map. It has 8925 training images and 1500 validation images, one for each image in Cityscapes for each intensity level. We evaluate on the full validation dataset.

\paragraph{BDD100k.}
The BDD100k \cite{yu2020bdd100k} dataset is another large autonomous driving dataset. Following previous works \cite{he2022cross,zhou2023unsupervised}, we train and evaluate on images of the Daytime split, which corresponds to a domain shift in camera and geographic location,  but without significant change in weather. It contains 36,728 training and 5258 validation images. It shares 7 of the 8 classes present in Cityscapes. 

\paragraph{ACDC.}
The Adverse Conditions Dataset with Correspondences (ACDC) \cite{sakaridis2021acdc} is an autonomous driving dataset focused on four adverse weather conditions: fog, night, rain and snow. It shares the same 8 classes as Cityscapes for object detection. Each split has 400 training and 100 validation images (106 for night validation). We train and evaluate on each of the individual splits. 

\subsection{Experimental Setup}\label{sec:experiment_setup}
We base our implementation on the Detectron2 framework \cite{wu2019detectron2}. For runs on Foggy Cityscapes and BDD100k, we follow past works \cite{cao2023contrastive} and use VGG16 \cite{simonyan2014very} pretrained on ImageNet \cite{deng2009imagenet} as the backbone. For runs on ACDC, we use the common ResNet-50 \cite{he2016deep}, also pretrained on ImageNet as the backbone. Single-scale Faster R-CNN is the detector in all experiments. Our models are trained on 2 RTX 6000 and use 8 source and 8 target images in each batch. We generally follow the test setting of Adaptive Teacher \cite{li2022cross}. When reimplementing baselines, images are scaled to have a height of 600 pixels for training. When aligning with DINO as per section \ref{sec:method_large_model_alignment}, we scale the height to 588 pixels to make it divisible by the ViT patch size of 14. All the results we present, including baselines, are evaluated from an EMA model $\bar{f}$ as discussed in \cref{sec:method_large_model_labels}.

In all of our runs and for all teachers, we use a pseudo-labelling threshold of ${\delta=0.8}$ and use weak-strong augmentations \cite{li2022cross} as our only augmentation. We set the loss hyperparameters in equation \eqref{eq:full_loss} as ${\lambda^{unsup}=1}$ and ${\lambda^{sim}=1}$. For runs on VGG16, we use a learning rate of 0.04 without decay, an EMA decay factor of ${\alpha=0.9996}$, pretrain on source only for ${n^{initPL}=20,\!000}$ iterations and train for a total of 60,000 iterations. For runs on ResNet-50, we use a learning rate of 0.01 without decay, an EMA decay factor of ${\alpha=0.999}$, pretrain on source for ${n^{initPL}=20,\!000}$ iterations and train for a total of 40,000 iterations.

We use the same DINOv2-based labeller to generate pseudo-labels for all runs following section \ref{sec:method_large_model_labels}. We train a Faster R-CNN head on top of a ViT-G encoder with frozen DINOv2 weights and use the same training setting and augmentations as the ResNet-50 student model but use only the source Cityscapes images. 

We use a frozen DINOv2 Vit-B as the alignment encoder following section \ref{sec:method_large_model_alignment}. The projection MLP has a hidden layer dimension of 1024. We initialize the similarity branch on source images until ${n^{initSim}=5000}$ iterations. We discuss the impact of the size of the DINO model in \cref{tab:ablat_sizes}. 

Because of the small size of the training and validation datasets for the individual ACDC splits, we present the average performance across three runs for both the baseline Adaptive Teacher and our method. When replicating the baseline, we use the same parameters used in \cite{li2022cross} (${\lambda_{unsup}=1}$, ${\lambda_{dis}=0.1}$) and start both the domain discrimination and the pseudo-labelling after 20,000 iterations. 

\begin{table*}[th!]
\centering
\begin{tabular}{c|c|ccccccc|c}
\hline Method & Type & Person & Rider & Car & Truck & Bus & Motor & Bicycle & mAP \\
\hline DA-Faster \cite{chen2018domain} & FR & 28.9 & 27.4 & 44.2 & 19.1 & 18.0 & 14.2 & 22.4 & 24.9 \\
SIGMA \cite{li2022sigma} & FCOS+GA & \cellcolor{YellowGreen!30} 46.9 & 29.6 & \cellcolor{YellowGreen!60} \underline{64.1} & 20.2 & 23.6 & 17.9 & 26.3 & 32.7 \\
TDD \cite{he2022cross} & FR+MT & 39.6 & 38.9 & 53.9 & 24.1 & 25.5 & \cellcolor{YellowGreen!30} 24.5 & 28.8 & 33.6 \\
PT \cite{chen2022learning} & FR+MT & 40.5 & 39.9 & 52.7 & 25.8 & \cellcolor{YellowGreen!30}33.8 & 23.0 & 28.8 & 34.9 \\
NSA \cite{zhou2023unsupervised} & FR+MT & - & - & - & - & - & - & - & 35.5 \\
REACT \cite{li2024react} & FR+MT & - & - & - & - & - & - & - & 35.8 \\
CAT \cite{kennerley2024cat} & FR+MT & 44.6 & \cellcolor{YellowGreen!60} \underline{41.5} & 61.2 & \cellcolor{YellowGreen!60} \underline{31.4} & \cellcolor{YellowGreen!60} \underline{34.6} & 24.4 & \cellcolor{YellowGreen!30} 31.7 & \cellcolor{YellowGreen!30} 38.5 \\
HT \cite{deng2023harmonious} & FCOS+MT & \cellcolor{YellowGreen!100} \textbf{53.4} & \cellcolor{YellowGreen!30} 40.4 & \cellcolor{YellowGreen!30}63.5 & \cellcolor{YellowGreen!30}27.4 & 30.6 & \cellcolor{YellowGreen!60} \underline{28.2} & \cellcolor{YellowGreen!60} \underline{38.0} & \cellcolor{YellowGreen!60} \underline{40.2} \\
\hline
DT (ours) & FR+MT & \cellcolor{YellowGreen!60}\underline{51.6} & \cellcolor{YellowGreen!100}\textbf{47.0} & \cellcolor{YellowGreen!100} \textbf{66.6} & \cellcolor{YellowGreen!100}\textbf{44.3} & \cellcolor{YellowGreen!100}\textbf{45.9} & \cellcolor{YellowGreen!100} \textbf{38.3} & \cellcolor{YellowGreen!100} \textbf{40.8} & \cellcolor{YellowGreen!100} \textbf{47.8}
\end{tabular}
\caption{\textbf{Results for domain adaptive object detection from Cityscapes to BDD100k.} Best performing methods are \textbf{bolded}, second best are \underline{underlined}.
}
\label{tab:res_BDD}
\end{table*}

\begin{table*}[ht!]
\centering
\begin{tabular}{c|c|cccccccc|c}
\hline Method & Type & Person & Rider & Car & Truck & Bus & Train & Motor & Bicycle & mAP \\
\hline DA-Faster \cite{chen2018domain} & FR & 29.2 & 40.4 & 43.4 & 19.7 & 38.3 & 28.5 & 23.7 & 32.7 & 32.0 \\
 DICN \cite{jiao2022dual} & FR & 47.3 & \cellcolor{YellowGreen!30} 57.4 & 64.0 & 22.7 & 45.6 & 29.6 & 38.6 & 47.4 & 44.1 \\
 NLTE \cite{liu2022towards} & FR+GA & 43.1 & 50.7 & 58.7 & 33.6 & 56.7 & 42.7 & 33.7 & 43.3 & 45.4 \\
 PT \cite{chen2022learning} & FR+MT & 43.2 & 52.4 & 63.4 & 33.4 & 56.6 & 37.8 & \cellcolor{YellowGreen!30} 41.3 & 48.7 & 47.1 \\
 MIC \cite{hoyer2023mic} & FR+MT & \cellcolor{YellowGreen!60} \underline{50.9} & 55.3 & \cellcolor{YellowGreen!60} \underline{67.0} & 33.9 & 52.4 & 33.7 & 40.6 & 47.5 & 47.6 \\
 AT \cite{li2022cross} & FR+MT & 45.5 & 55.1 & 64.2 & 35.0 & 56.3 & \cellcolor{YellowGreen!100}\textbf{54.3} & 38.5 & 51.9 & 50.9 \\
 CMT \cite{cao2023contrastive} & FR+MT & 47.0 & 55.7 & 64.5 & \cellcolor{YellowGreen!60} \underline{39.4} & \cellcolor{YellowGreen!60} \underline{63.2} & 51.9 & 40.3 & \cellcolor{YellowGreen!30} 53.1 & \cellcolor{YellowGreen!30} 51.9 \\
 REACT \cite{li2024react} & FR+MT & \cellcolor{YellowGreen!100} \textbf{51.4} & \cellcolor{YellowGreen!60} \underline{57.9} & \cellcolor{YellowGreen!100} \textbf{67.4} & \cellcolor{YellowGreen!30} 37.7 & \cellcolor{YellowGreen!30} 58.4 & \cellcolor{YellowGreen!30} 52.8 & \cellcolor{YellowGreen!60} \underline{44.6} & \cellcolor{YellowGreen!60} 54.6 & \cellcolor{YellowGreen!60} 53.1 \\
\hline
DT (ours) & FR+MT & \cellcolor{YellowGreen!30} 48.5 & \cellcolor{YellowGreen!100} \textbf{60.0} & \cellcolor{YellowGreen!30}65.4 & \cellcolor{YellowGreen!100} \textbf{47.2} & \cellcolor{YellowGreen!100} \textbf{66.5} & \cellcolor{YellowGreen!60} \underline{52.9} & \cellcolor{YellowGreen!100} \textbf{46.2} & \cellcolor{YellowGreen!100} \textbf{56.7} & \cellcolor{YellowGreen!100} \textbf{55.4}
\end{tabular}
\caption{\textbf{Results for domain adaptive object detection from Cityscapes to Foggy Cityscapes for the entire validation set.} Best performing methods are \textbf{bolded}, second best are \underline{underlined}.
}
\label{tab:res_cityfog_all}
\end{table*}

\subsection{Results}
\label{sec:results}
We compare the performance of our method with other state-of-the-art methods and report class-averaged precision at $50\%$ IoU (mAP@50). In the tables, we use the following abbreviations: $\textbf{FR}$ for Faster R-CNN, $\textbf{GA}$ for graph alignment, and $\textbf{MT}$ for Mean Teacher. We evaluate across different environments, cameras and adverse weather conditions.

\paragraph{Cityscapes $\rightarrow$ BDD100k.}%
The first test setting considers domain adaptation to BBD100K, corresponding to different environment and camera. We present our results in \cref{tab:res_BDD}. We obtain state-of-the-art results, exceeding the performance of Harmonious Teacher \cite{deng2023harmonious} by $+7.6\%$, with particularly large improvements on the rarer classes. Given a larger target dataset, we show that improving the quality of labels by using a better labeller improves performance substantially over using the Mean Teacher framework. We discuss the correlation between labeller size and student network performance in \cref{tab:label_size}.


\paragraph{Cityscapes $\rightarrow$ Foggy Cityscapes.}
The second test considers domain adaptation across a relatively easy adverse weather condition, the synthetic addition of fog. We present our results in \cref{tab:res_cityfog_all} for the full validation set and obtain state-of-the-art performance on this dataset, improving over the recent REACT \cite{li2024react} method by $+2.3\%$ and our baseline architecture AT by $+4.5\%$ when using the ViT-G labels. 
Compared to our results on BDD100k, we show more minor improvements on this dataset. We believe this is due to the small domain gap between Cityscapes and Foggy Cityscapes, which is generated by a consistent procedure on the original Cityscapes images and uses the same labels.


\paragraph{Cityscapes $\rightarrow$ ACDC splits.}
The final test setting considers domain adaptation in the most complex setting: adverse weather, different camera and environment, and with a much smaller target dataset. We present our results in \cref{tab:res_acdc}. Here as well, we observe consistent and significant performance improvements compared to the Adaptive Teacher baseline, improving by $+6.4\%$ on fog, $+6.9\%$ on night, $+1.3\%$ on rain and $+1.6\%$ on snow. While the DINO pseudo-labels do better than Mean Teacher labels, particularly for the most challenging night split, we believe the quality of pseudo-labels limits our improvements. We discuss this further in section \ref{sec:dino_vs_student} below. Another issue is the small size of the target dataset, which leads to overfitting on the pseudo-labels and restricts performance on the test set.

\begin{table}[t]
\centering
\begin{tabular}{c|c|c|c|c}
\hline
Method & Fog & Night & Rain & Snow \\
\hline AT$^{\dagger}$ \cite{li2022cross} & 62.2 & 29.5 & 37.7 & 55.2 \\
DT-G (ours) & 68.6 & 36.4 & 39.0 & 56.8 \\
\end{tabular}
\caption{\textbf{Results for domain adaptive object detection from Cityscapes to the 4 ACDC splits.} \textsuperscript{$\dagger$}We reimplement AT from publicly available code.}
\label{tab:res_acdc}
\end{table}



\subsection{Ablations} \label{sec:ablations}
We run our ablations on the Cityscapes to BDD100k test case, as the large dataset size and significant domain gap best exemplify performance improvements from different methodological choices.

\paragraph{Ablations on Components.}
\cref{tab:ablat_components2} presents the ablation on the contributions of each component of our approach. Starting from the Adaptive Teacher baseline, we substitute the standard EMA teacher labeller \textbf{MT} with our DINO labeller \textbf{DL} (section \ref{sec:method_large_model_labels}) and substitute the domain adversarial loss $\mathcal{L}^{dis}$ with our patch similarity loss $\mathcal{L}^{sim}$ (section \ref{sec:method_large_model_alignment}). We report the best performance and performance at $n^{initPL}$ iterations, just before any pseudo-labels are used, to separate the impacts of changes in alignment and labelling methods. We see that while our alignment loss $\mathcal{L}^{sim}$ improves performance by $+4.0\%$ before pseudo-labels are introduced and overall by $+3.5\%$ (AT vs case 1), it doesn't allow the Mean Teacher framework to generate improvements similar to the DINO labeller (case 1 vs DT). Using DINO labels improves performance by over $10\%$ regardless of the type of alignment used (AT vs case 2 or DT).



\begin{table}[t]
\centering
\begin{tabular}{c|cc|cc}
\hline
Method & Label. & Align. &  $n^{initPL}$ & Best mAP \\
\hline AT$^{\dagger}$ \cite{li2022cross} & \textbf{MT} & $\mathcal{L}^{dis}$ & 28.5 & 31.8 \\
case 1 & \textbf{MT} & $\mathcal{L}^{sim}$ & 32.5 & 35.3 \\
case 2 & \textbf{DL} & $\mathcal{L}^{dis}$ & 28.5 & 46.8\\
DT & \textbf{DL} & $\mathcal{L}^{sim}$ & 33.0 & 47.8 \\
\end{tabular}
\caption{\textbf{Ablation of the DT components on Cityscapes to BDD100k.}  We use \textbf{MT} for Mean Teacher labels and \textbf{DL} for DINO labeller. \textsuperscript{$\dagger$}We reimplement AT from publicly available code.}
\label{tab:ablat_components2}
\end{table}

\paragraph{Teacher Size.}
Next, we consider the impact of the size of the DINOv2 model used as the backbone to generate pseudo-labels (section \ref{sec:method_large_model_labels}) and as the alignment target (section \ref{sec:method_large_model_alignment}). Results are presented in \cref{tab:ablat_sizes}. The ViT-S, B, L and G models have 21, 86, 300 and 1,100 million parameters respectively. \cref{tab:label_size} presents the ablation on DINO labeller size, and scaling the backbone from ViT-B to ViT-G improves our results by $+1.8\%$. 
Interestingly, the final student model is better than the labeller for ViT-B and ViT-L, meaning labeller performance is not a direct upper bound. While the labeller is trained on source only, the student uses a high-confidence subset of target labels and true source labels to learn a representation that generalizes to both domains. From \cref{tab:align_size}, we also find that aligning with the larger ViT-B model improves by $+0.4\%$ compared to ViT-S, and both do better than the baseline. 


\begin{table}[t]
    \begin{subtable}{.46\linewidth}\centering
        {\begin{tabular}{c|cc}
        \hline
        Method & Lab. & Stud. \\
        \hline
        \textbf{MT}   & -    & 35.3 \\
        \textbf{DL-B} & 42.7 & 46.0 \\
        \textbf{DL-L} & 45.7 & 46.9 \\
        \textbf{DL-G} & 51.1 & 47.8 \\
        \end{tabular}}
        \caption{Labeller size}\label{tab:label_size}
    \end{subtable}%
    \hspace{2pt}
    \begin{subtable}{.51\linewidth}\centering
        {\begin{tabular}{c|cc}
        \hline
        Method & $n^{initPL}$ & mAP  \\
        \hline
        $\mathcal{L}^{dis}$ & 28.5 & 46.8 \\
        $\mathcal{L}^{sim}\mbf{S}$ & 31.9 & 47.4 \\
        $\mathcal{L}^{sim}\mbf{B}$ & 33.0 & 47.8 \\
        \end{tabular}}
        \vspace{0.1in}
        \caption{Alignment teacher size}\label{tab:align_size}
    \end{subtable}
\caption{\textbf{Ablations of the size of DINOv2 backbone model for the labeller and alignment teacher.} We consider ViT-B, L and G for the DINO labeller (\textbf{DL-B}, \textbf{DL-L}, \textbf{DL-G}), and ViT-S and B for the alignment loss ($\mathcal{L}^{sim}\mbf{S}$, $\mathcal{L}^{sim}\mbf{B}$).}
\label{tab:ablat_sizes}
\end{table}


\subsection{DINO Teacher vs Other Baselines} \label{sec:dino_vs_student}
Finally, we consider how our DINO Teacher compares to other strategies for object detection on the target dataset and present the results in \cref{tab:student_vs_teacher}. First, the ViT-G DINO labeller generalizes much better than the source-only student model, validating our choice of using a VFM labeller for pseudo-label generation.
Our DT approach significantly outperforms source-only student training and target-only training on the small ACDC splits, but it lags behind target-only training on the larger BDD dataset, showing the advantages of domain adaptation for small datasets. An alternative upper bound for domain adaptation is the performance with an ideal labeller, where ground-truth target domain labels are available and training is done on source and target images, denoted w/ GT labels in \cref{tab:student_vs_teacher}. This improves performance compared to target-only training on ACDC but not for BDD, meaning training with the source data can reduce performance given sufficient labels on the larger target dataset. Our DT model is close to this bound, but the gap indicates that a better labeller could yield improvements.

\begin{table}[t]
\centering
\begin{tabular}{c|c|cccc}
\hline
\multirow{2}{*}{Method} & \multirow{2}{*}{BDD} & \multicolumn{4}{c}{ACDC} \\
& & Fog & Night & Rain & Snow \\
\hline
Source only  & 29.1 & 61.2 & 19.7 & 37.0 & 46.3 \\
Target only  & 54.4 & 48.6 & 24.6 & 30.5 & 46.0 \\
ViT-G labeller & 51.1 & 65.8 & 39.4 & 47.3  & 49.3 \\
w/ GT labels$^\dagger$  & 51.6 & 70.3 & 37.2 & 44.0 & 57.5  \\
\hline DT$^{*}$ & 47.8 & 68.6 & 36.4 & 39.0 & 56.8 \\
\end{tabular}
\caption{\textbf{Performance of the DT model compared to other strategies.} \textsuperscript{$*$}Reported results for DT on ACDC are the average of 3 runs. \textsuperscript{$\dagger$}Network trained on source and target with all labels.}
\label{tab:student_vs_teacher}
\end{table}

\section{Conclusion}
\label{sec:conclusion}
We present a novel framework for domain adaptive object detection, DINO Teacher, in which we leverage the generalizability of the features of vision foundational models such as DINOv2. Our DINO Teacher approach is composed of two components. First, we create a new labeller by training a detection head on a frozen VFM backbone using only source data. We show that this simple framework outperforms the standard Mean Teacher self-labelling approach, particularly on the BDD100k dataset, demonstrating the limitations of the standard Mean Teacher architecture. Next, we show that aligning image features from the backbone of the student model with those from a frozen VFM encoder independently for source and target images pushes the student to learn features that are more discriminative and generalize better from the source to the target domain, improving performance.
\clearpage
\setcounter{page}{1}
\setcounter{table}{6}
\setcounter{figure}{3}
\maketitlesupplementary
\appendix

\section{Algorithm and Implementation Details}
\label{sec:pseudocode}
Our proposed algorithm trains the student model in multiple phases for both target domain pseudo-label training and alignment. We present all the training phases here and summarize our method in Algorithm \ref{alg:dino_align}. For pseudo-label training, we have two phases:
\begin{packedEnum}
    \item If $iter < n^{initPL}$, train only on source images with ground truth labels.
    \item If $iter \geq n^{initPL}$, train on source images with ground truth labels and on target images with DINO labeller labels.
\end{packedEnum}
In all of the presented runs, we use ${n^{initPL}=20,\!000}$. For alignment, we have 2 phases:
\begin{packedEnum}
    \item If $iter < n^{initSim}$, align student and DINO features only from source images.
    \item If $iter \geq  n^{initSim}$, align student and DINO features from both source and target images.
\end{packedEnum}
In all of the presented runs, we use ${n^{initSim}=5000}$. With weak-strong augmentation \cite{liu2021unbiased}, two versions of each source and target image are generated. We do labelled detection training and DINO alignment with both weakly and strongly augmented source images and target images. We present Algorithm \ref{alg:dino_align} assuming that ${n^{initSim}\leq n^{initPL}}$, which is true for all the reported runs.

\begin{algorithm}
\caption{Pseudocode for DINO Teacher}\label{alg:dino_align}
\begin{algorithmic}
\While{$iter < n^{max}$}
    \State \textbf{Input:} Source images and labels $\left(X_{S},Y_{S},B_{S}\right)$
    \State \textbf{Input:} Target images $X_{T}$ 
    \State Augment source images $\mathbf{X}_{S} = \left[X_{S}^{weak},X_{S}^{strong}\right]$
    \State Compute source similarity loss $\mathcal{L}^{sim}_{S}\left(\mathbf{X}_{S}\right)$
    \State Compute source detection loss $\mathcal{L}^{det}_{S}\left(\mathbf{X}_{S},Y_{S},B_{S}\right)$
    \If{$iter \geq n^{initSim}$}
        \State Augment target images $\mathbf{X}_{T} = \left[X_{T}^{weak},X_{T}^{strong}\right]$
        \State Compute target similarity loss $\mathcal{L}^{sim}_{T}\left(\mathbf{X}_{T}\right)$
    \EndIf
    \If{$iter \geq n^{initPL}$}
        \State Get DINO labels $\left(\tilde{Y}_{T},\tilde{B}_{T}\right)$
        \State Compute target detect. loss $\mathcal{L}^{det}_{T}\left(\mathbf{X}_{T},\tilde{Y}_{T},\tilde{B}_{T} \right)$
    \EndIf

\State Sum all losses and update student model
\EndWhile

\end{algorithmic}
\end{algorithm}

\section{Additional Ablations and Results}
\subsection{Alignment with Larger ViT-L} 
In our ablations in the main paper in Tab. 5b, we present results for alignment with a ViT-S and ViT-B DINOv2 backbone and discuss in Sec. 3.3 that using larger models during online training was too time consuming. We add the ViT-L result in \cref{tab:align_size_full} and show the continuing trend of larger alignment targets leading to improvements in final performance. This shows the potential of aligning with larger models if the training cost could be reduced. One possible solution could be to precompute and store the DINO features for each image in the target dataset, thus only requiring a single forward pass on the dataset for the alignment target. This single forward pass could even be combined with the forward pass of the DINO labeller used to generate the target pseudo-labels. However, this would have significant data storage requirements for the larger datasets.
\begin{table}[t]
    \centering
    {\begin{tabular}{c|cc}
        \hline
        Method & $n^{initPL}$ & mAP  \\
        \hline
        $\mathcal{L}^{dis}$ & 28.5 & 46.8 \\
        $\mathcal{L}^{sim}\mbf{S}$ & 31.9 & 47.4 \\
        $\mathcal{L}^{sim}\mbf{B}$ & 33.0 & 47.8 \\
        $\mathcal{L}^{sim}\mbf{L}$ & 32.3 & 48.3 \\
    \end{tabular}}
    \caption{\textbf{Ablation of the size of DINOv2 backbone model for the alignment teacher.} We consider ViT-S, B and L for the alignment loss ($\mathcal{L}^{sim}\mbf{S}$, $\mathcal{L}^{sim}\mbf{B}$, $\mathcal{L}^{sim}\mbf{L}$).}
    \label{tab:align_size_full}
\end{table}

\subsection{Ablation on the Choice of Student Backbone}\label{sec:choice_bbone}
Our main results in Sec. 4.3 use VGG16 as the backbone for domain adaptation to Foggy Cityscapes and to BDD100k, following previous works. However, we chose ResNet-50 for our experiments on ACDC as this is a more common architecture. Similarly, the small ViT-S architecture is of similar size to VGG and is generally a stronger baseline, particularly when using better initializations such as the distilled DINOv2 weights.

We present additional results on the Cityscapes $\rightarrow$ BDD100k test case with a ResNet-50 backbone and a ViT-S backbone for 3 settings: source only (SO), self-generated pseudo labels with Mean Teacher (MT) and our DINO Teacher (DT) in \cref{tab:bdd_r50}. We use the same training protocol for ResNet-50 as our tests on ACDC described in Sec. 4.2 and use the same ViT-G generated labels on BDD100k as in our original test. We consider cases where the ViT-S backbone is frozen and unfrozen. When unfrozen, the ViT-S backbone learning rate is scaled down by a factor of 0.01 compared to the detector head learning rate. 

For all settings, the better architectures lead to improved performance compared to the VGG results. Moreover, our DINO Teacher remains significantly better ($+5\%$) compared to Mean Teacher for all tested architectures. Even with the stronger student models, we see large improvements when using labels from the external labeller instead of self-labelling. 
\begin{table}[t]
    \centering
    {\begin{tabular}{ccc|c}
        \hline
        Backbone & State & Method & mAP  \\
        \hline
        \multirow{3}{*}{VGG16} & \multirow{3}{*}{Unfr.} & SO & 29.1 \\
        & & MT & 30.1 \\
        & & DT & 47.8 \\
        \hline
        \multirow{3}{*}{ResNet-50} & \multirow{3}{*}{Unfr.} & SO & 37.9 \\
        & & MT & 42.7 \\
        & & DT & 52.5 \\
        \hline
        \multirow{3}{*}{ViT-S} & \multirow{3}{*}{Frozen} & SO & 37.3 \\
        & & MT & 38.2 \\
        & & DT & 43.2 \\
        \hline
        \multirow{3}{*}{ViT-S} & \multirow{3}{*}{Unfr.} & SO & 43.9 \\
        & & MT & 47.1 \\
        & & DT & 53.9 \\
    \end{tabular}}
    \caption{\textbf{Ablation of the choice of student backbone model.} We maintain significant improvements on ResNet-50 and ViT-S.}
    \label{tab:bdd_r50}
\end{table}

\subsection{Ablation on Choice of Detector}
We use single-scale (SS) Faster R-CNN in all our results, but some works used multi-scale Faster R-CNN (FR) with a feature pyramid (FPN) or single-stage methods like FCOS. We provide results for both in \cref{tab:multiscale}, comparing source only (SO) and the self-generated pseudo-labels from Mean Teacher (MT) to our DINO Teacher (DT). Again, we see that our DINO Teacher approach gives substantial improvements for all tested detectors. Moreover, there is a larger improvement ($+1.0\%$ to $+2.8\%$) when going from single-scale to multi-scale Faster R-CNN for Mean Teacher compared to source only, highlighting that the performance of Mean Teacher self-labelling approaches is sensitive to the source-only model performance from which they are initialized. We observe similar trends when considering the changes in backbones in \cref{tab:bdd_r50}. We are unable to generate good Mean Teacher pseudo-labels for the single-stage FCOS approach, and find that the box confidence scores are generally much lower than those of the two-stage Faster R-CNN methods. We test multiple bounding box confidence thresholds from $0.3$ to $0.8$ for pseudo-label selection but all led to rapid performance drop on both source and target datasets, meaning best performance occurs prior to using any pseudo-labels.

\begin{table}[t]
    \centering
    {\begin{tabular}{ccc|c}
        \hline
        Detector & Features & Method & mAP  \\
        \hline
        \multirow{3}{*}{FR} & \multirow{3}{*}{SS} & SO & 29.1 \\
        & & MT & 30.1 \\
        & & DT & 47.8 \\
        \hline
        \multirow{3}{*}{FR} & \multirow{3}{*}{FPN} & SO & 32.7 \\
        & & MT & 35.5 \\
        & & DT & 51.2 \\
        \hline
        \multirow{3}{*}{FCOS} & \multirow{3}{*}{FPN} & SO & 33.1 \\
        & & MT & 33.1$^{*}$ \\
        & & DT & 51.8 \\
    \end{tabular}}
    \caption{\textbf{Ablation of the choice of student model detector.} We maintain significant improvements across detectors. \textsuperscript{$*$}We do not obtain good self-generated pseudo-labels with MT on FCOS, and performance collapses rapidly when using the generated labels.}
    \label{tab:multiscale}
\end{table}

\subsection{Ablation on Unfreezing the Labeller Backbone}
In the main paper, we consider the simplest setting in which the labeller ViT backbone is kept frozen, maintaining the pretrained DINOv2 weights, and present ablations on the effects of labeller backbone size and performance in Tab. 5a. Here, we consider the case where the labeller backbone is unfrozen and is trained with the detector on source images.  We generally follow the training regiment from Sec. 4.2 when training, but downscale the learning rate of the backbone by a factor of 0.01. We present the performance of the labeller trained on Cityscapes images on the unseen target BDD and the performance of the VGG student trained with the labeller pseudo-labels on BDD in \cref{tab:unfreeze_bbone}. We see that while unfreezing the backbone leads to significant improvements on the labeller ($+4.3\%$), these are smaller for the student ($+0.3\%$). We find that the improvements in labeller performance due to unfreezing do not transfer to the student as well as those from using a larger model.

\begin{table}[t]
    \centering
    {\begin{tabular}{cc|cc}
        \hline
        Backbone & State & Labeller & Student  \\
        \hline
        \multirow{2}{*}{ViT-L} & Frozen & 45.7  & 46.9 \\
        & Unfr. & 50.0 & 47.2 \\ \hline
        ViT-G & Frozen & 51.1 & 47.8 
    \end{tabular}}
    \caption{\textbf{Ablation on unfreezing the labeller backbone.} Performance of the labeller and the student improve when unfrozen, but the transfer is less effective compared to using a larger frozen backbone.}
    \label{tab:unfreeze_bbone}
\end{table}
\subsection{Ablation on Using EMA Teacher Labels} 
\label{sec:abl_EMA}
Our method differs from the Mean Teacher baseline by generating target pseudo-labels with our DINO labeller instead of using the EMA teacher, leading to better performance. However, in many cases (smaller labeller in Tab. 5b, ACDC fog and snow splits, larger student backbone and detectors), the teacher model derived from the student exposed to target data performed better than the labeller trained only on source data. Thus, we investigate whether EMA teacher labels could be useful in addition to using the DINO labeller. We assume that one of the most significant advantages of the DINO labels is that they are more accurate for the first few iterations of training on the target data, after which the EMA teacher is adapted to the target domain and could be good enough to generate useful labels.

We propose to add a third phase to pseudo-label training (see \cref{sec:pseudocode} above): after $n^{initEMA}$ iterations, we start using EMA teacher pseudo-labels following the regular Mean Teacher strategy. We propose two variations: use only EMA teacher labels after $n^{initEMA}$ (EMA only), or alternate between DINO ViT-G and EMA labels in alternating iterations (EMA mixed). The second variation assumes that there might be some advantage in using the EMA labels but tries to avoid any potential issue of drift from biased EMA teacher labels by still using the DINO labels.

We present results in \cref{tab:EMA_teacher} of these variations on transfer to BDD with a VGG backbone and compare to the nominal DINO Teacher (DT) and Mean Teacher with DINO alignment but no labels (MT, corresponds to case 1 in Tab. 4). We follow the training protocol from Sec. 4.2 and use ${n^{initEMA}=25,\!000}$. For both variations, the best performance was worse than the nominal DT but better than MT. Notably, the EMA only case led to a monotonic reduction in performance from the initial state at ${n^{initEMA}=25,\!000}$ iterations, meaning the EMA pseudo-labels made performance worse after the initial training phase with DINO labels. This highlights the potential risk of using pseudo-labels and the importance of more robust generation. 

\begin{table}[t]
    \centering
    {\begin{tabular}{c|c}
        \hline
        Method & mAP  \\
        \hline
        MT & 35.3 \\
        EMA only & 43.3$^*$ \\
        EMA mixed & 47.3 \\
        \hline
        DT & 47.8 \\
    \end{tabular}}
    \caption{\textbf{Ablation of using EMA teacher pseudo-labels.} \textsuperscript{$*$}Best performance occurred at 25,000 iterations, just before switching to using only EMA teacher pseudo-labels.}
    \label{tab:EMA_teacher}
\end{table}

\subsection{Issues with Self-Labelling Approaches}
When looking at our results comparing self-generated labels with using an external source of labels, we generally find that performance is not only worse when self-labeling but it can be unstable: after some initial improvements compared to source-only training, performance on the target data can drop. This is most striking in our ablations in \cref{tab:EMA_teacher}, where even when initialized with the stronger DINO Teacher labels, using only Mean Teacher labels after ${n^{initEMA}}$ leads to a reduction in performance over time. We believe this can be explained by class confusion in the pseudo-labels, particularly for rare classes, which lead to degraded class representation over time. 

\begin{figure*}[t]
  \centering
    \includegraphics[width=0.90\textwidth]
    {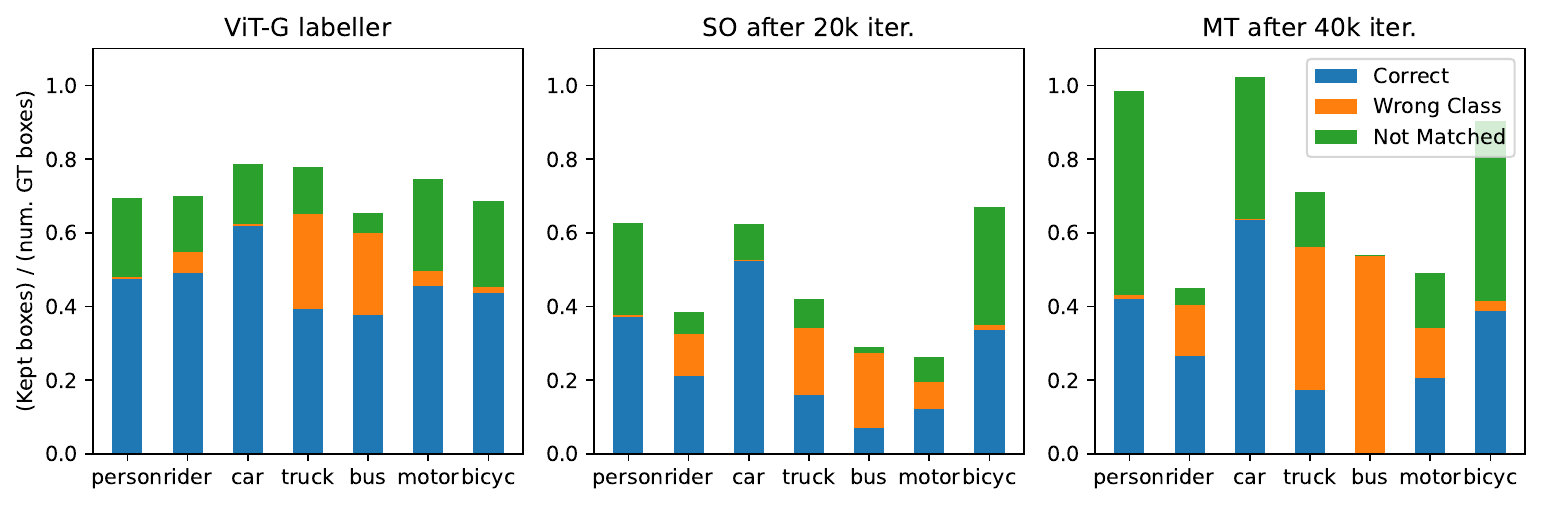}
    	\caption{\textbf{Quality of generated pseudo-labels.} Ratio of number of high-confidence pseudo-labels compared to the total number of instances per class. The student model (SO and MT) is much weaker for the rare classes, and as training progresses Mean Teacher pseudo-labels, the label quality becomes worse.}
    	\label{fig:quality_pl}
\end{figure*}

To explore this, \cref{fig:quality_pl} presents the ratio of pseudo-labels with class confidence values above the threshold of ${\delta=0.8}$ (and thus kept as labels) compared to the number of real instances of a given class on the target BDD dataset for three labellers: our ViT-G DINO labeller, source-only VGG student at ${n^{initPL}=20,\!000}$ iterations (when Mean Teacher pseudo-label generation begins), and the Mean Teacher EMA teacher at $40,\!000$ iterations. We observe that the source-only student is a poor labeller for rare classes like Truck, Bus and Motorcycle, and this poor initial performance leads to even worse labels as training continues. We see an increase in confident boxes of common classes like car that are matched to rare class boxes (orange lines for truck and bus) or that do not match any ground truth instance (green bar on car).

\subsection{Results on BDD Daytime-Sunny to BDD Night-Sunny}
We followed existing domain adaptation works in considering domain adaptation from Cityscapes to BDD100k Daytime in our main results in Tab. 1. However, the BDD100k \cite{yu2020bdd100k} dataset contains many images that are not daytime scenes, including many night scenes. A recent work in domain generalization by \citet{wu2022single} proposes a new dataset of multiple weather conditions composed mainly of BDD100k images. It contains five splits: daytime-sunny (day and without significant weather) from BDD100k, night-sunny also from BDD100k, the smaller dusk-rainy and night-rainy rendered from BDD100k images, and daytime-foggy composed of Foggy Cityscapes \cite{sakaridis2018semantic} and Adverse-Weather \cite{hassaballah2020vehicle} images. The proposed daytime-sunny split differs from the Daytime split we present in Sec. 4.1. 

We consider the daytime-sunny and night-sunny splits, which have a similar number of images, with 19,395 training and 8313 testing images for daytime-sunny and 18,310 training and 7848 testing images for night-sunny. This test case is similar to Cityscapes $\rightarrow$ Foggy Cityscapes except for larger datasets and a non-synthetic domain gap. We use the training protocol described in Sec. 4.2 for a VGG16 backbone. Because we use a new source dataset, we train a new ViT-L DINO labeller on the daytime-sunny split. We present results for the Adaptive Teacher baseline (AT) and our DINO Teacher (DT) in \cref{tab:bdd_ds2ns}. Similar to the domain adaptation from Cityscapes to BDD100k Daytime, we find that DINO teacher leads to a substantial improvement of $+9.2\%$ on the target dataset compared to Adaptive Teacher.

\begin{table}[t]
    \centering
    {\begin{tabular}{c|c}
        \hline
        Method & mAP  \\
        \hline
        AT & 34.9 \\
        DT & 44.1 \\
    \end{tabular}}
    \caption{\textbf{Results for domain adaptive object detection from BDD daytime-sunny to BDD night-sunny.}  We maintain significant improvements on this task.}
    \label{tab:bdd_ds2ns}
\end{table}


\subsection{Complete ACDC results} 
\label{sec:acdc_results}
We present the full per-class AP and mAP for the ACDC runs that are averaged to obtain the results of Tab. 3 in the paper. \cref{tab:res_acdc_AT_all} presents results from our reimplementation of Adaptive Teacher, and \cref{tab:res_acdc_DT_all} presents our DINO teacher results. We show that our DINO Teacher consistently improves on the Adaptive Teacher baseline. However, both the baseline and our proposed method struggle with certain rare classes in the hardest night and rain splits, specifically Rider and Bicycle. This is not seen on the easier fog and snow splits or when adapting to BDD100k Daytime (Tab. 1) or to Foggy Cityscapes (Tab. 2). This could be because of limited labels, overlapping boxes between Rider and Bicycle instances, or Rider instances being incorrectly pseudo-labelled as Person, all of which could cause training issues.

\section{Qualitative Results}
\label{sec:visuals}
\begin{figure*}[ht]
    \centering
    \stackunder[5pt]{\includegraphics[width=0.35\textwidth]{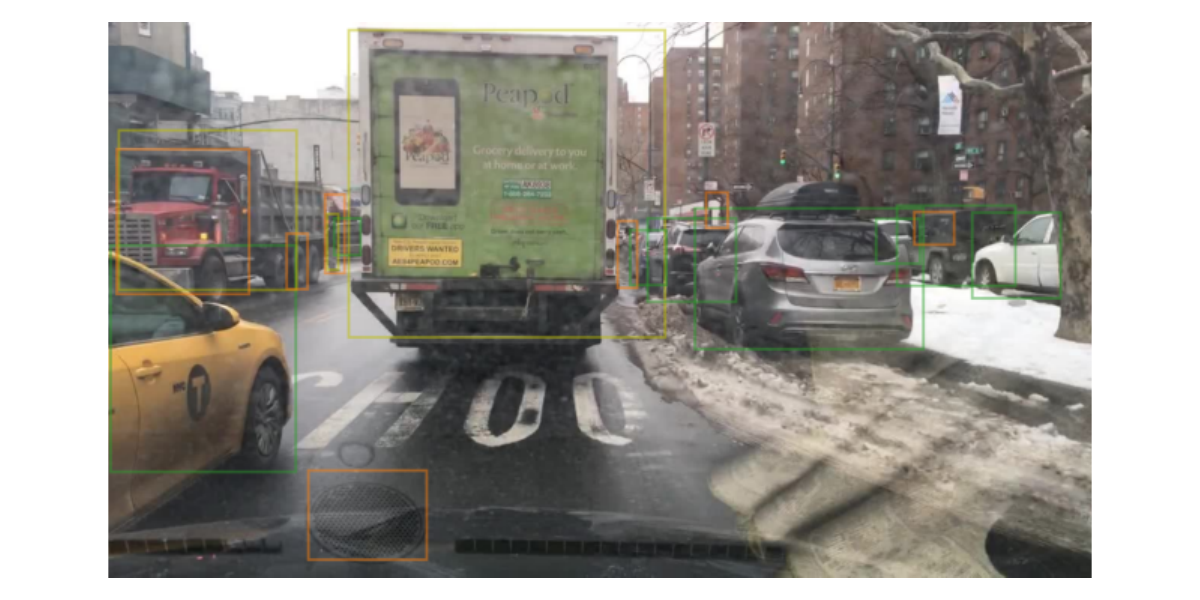}}{}
    \stackunder[5pt]{\includegraphics[width=0.35\textwidth]{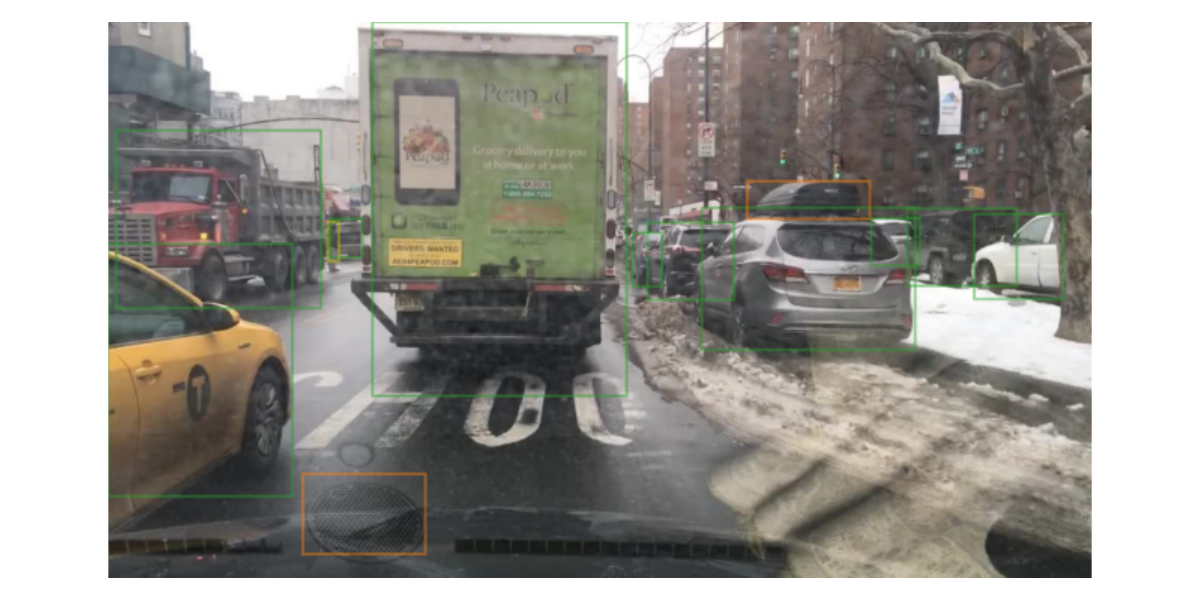}}{}
    \stackunder[5pt]{\includegraphics[width=0.35\textwidth]{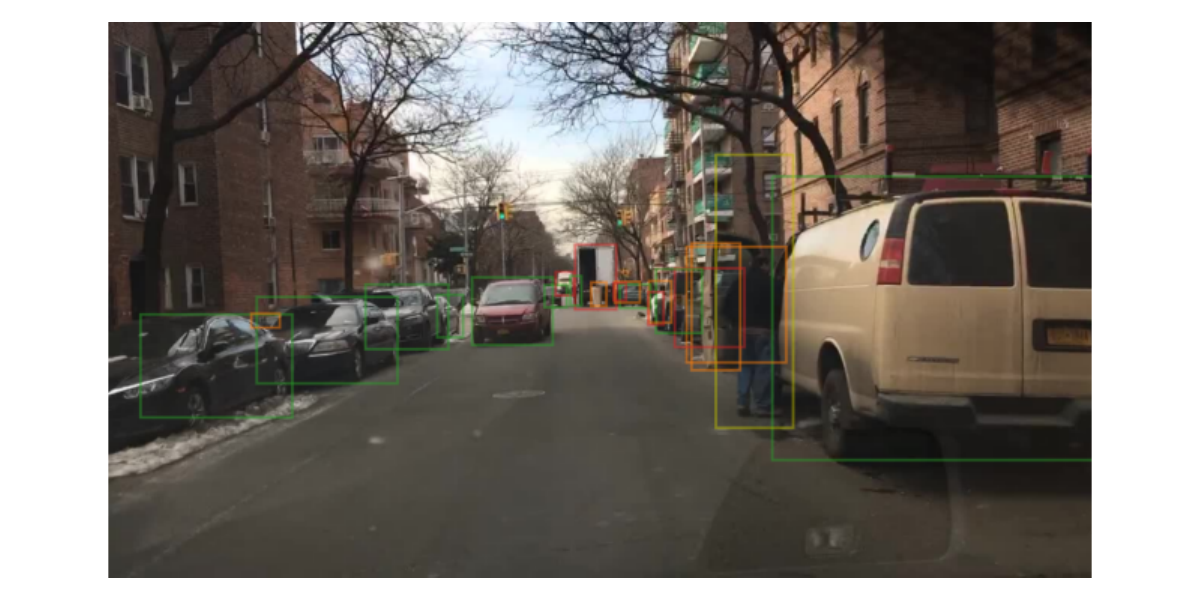}}{}
    \stackunder[5pt]{\includegraphics[width=0.35\textwidth]{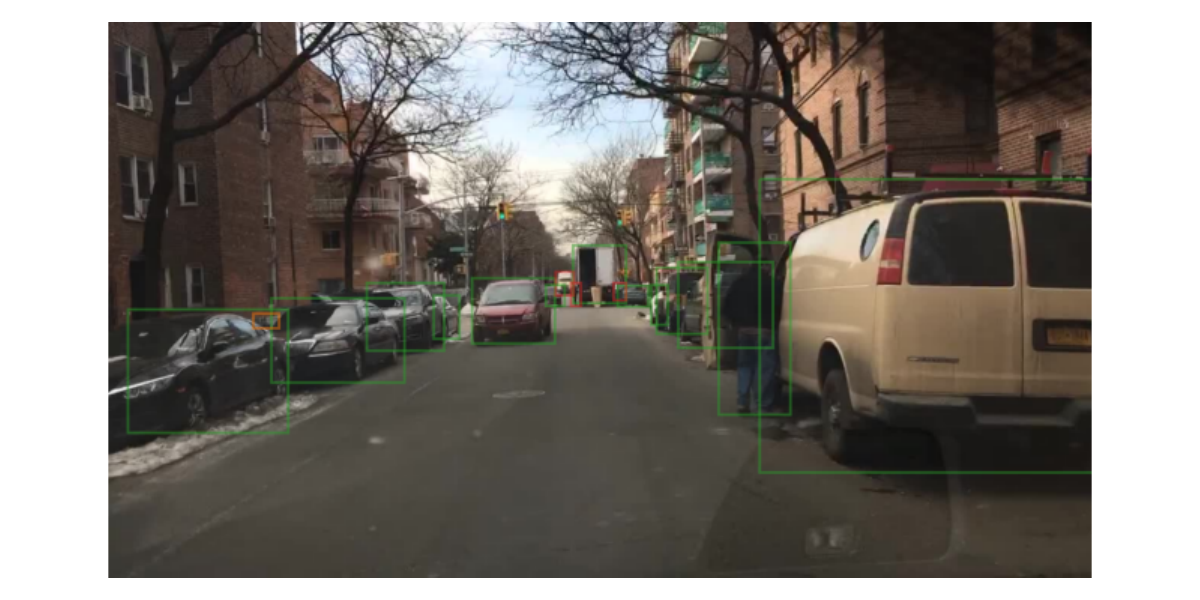}}{}
    \stackunder[5pt]{\includegraphics[width=0.35\textwidth]{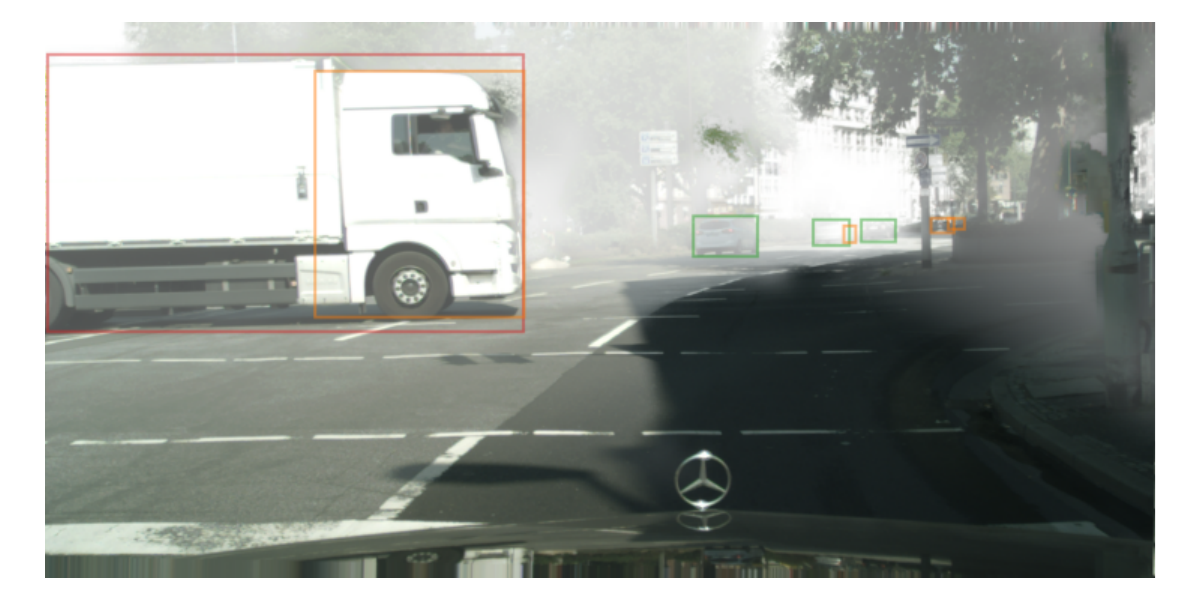}}{}
    \stackunder[5pt]{\includegraphics[width=0.35\textwidth]{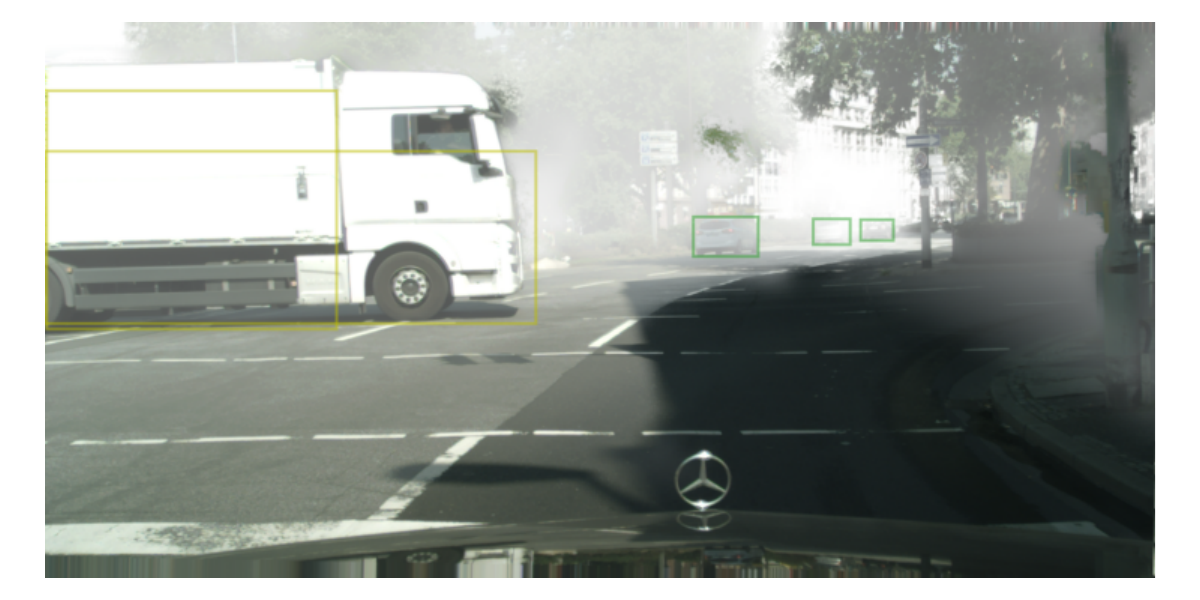}}{}
    \stackunder[5pt]{\includegraphics[width=0.35\textwidth]{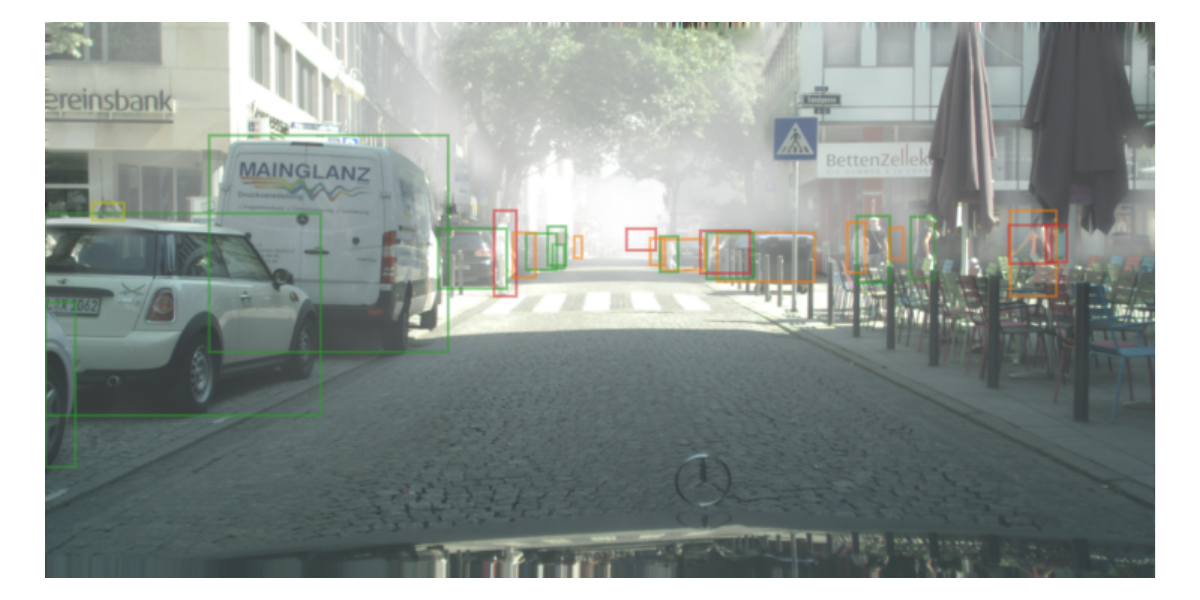}}{}
    \stackunder[5pt]{\includegraphics[width=0.35\textwidth]{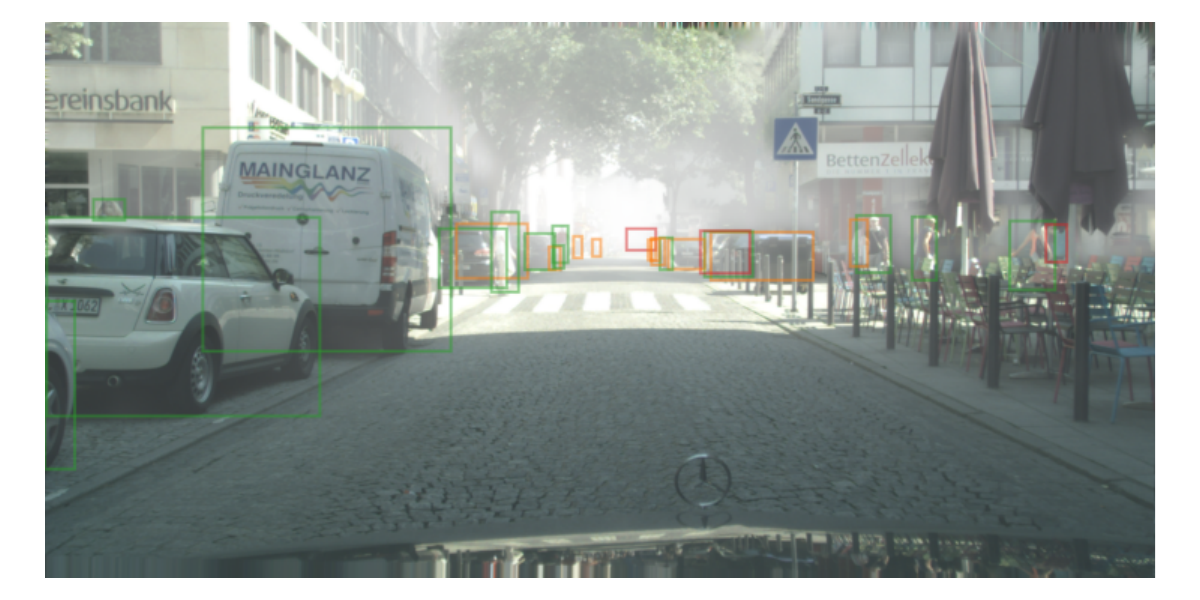}}{}
    \stackunder[5pt]{\includegraphics[width=0.35\textwidth]{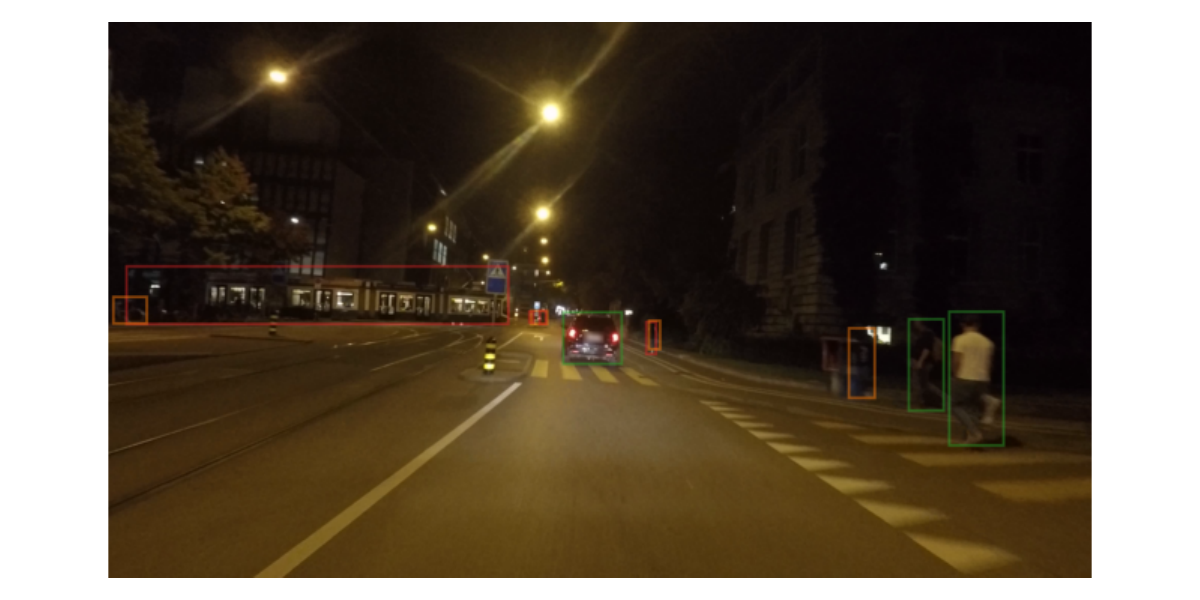}}{}
    \stackunder[5pt]{\includegraphics[width=0.35\textwidth]{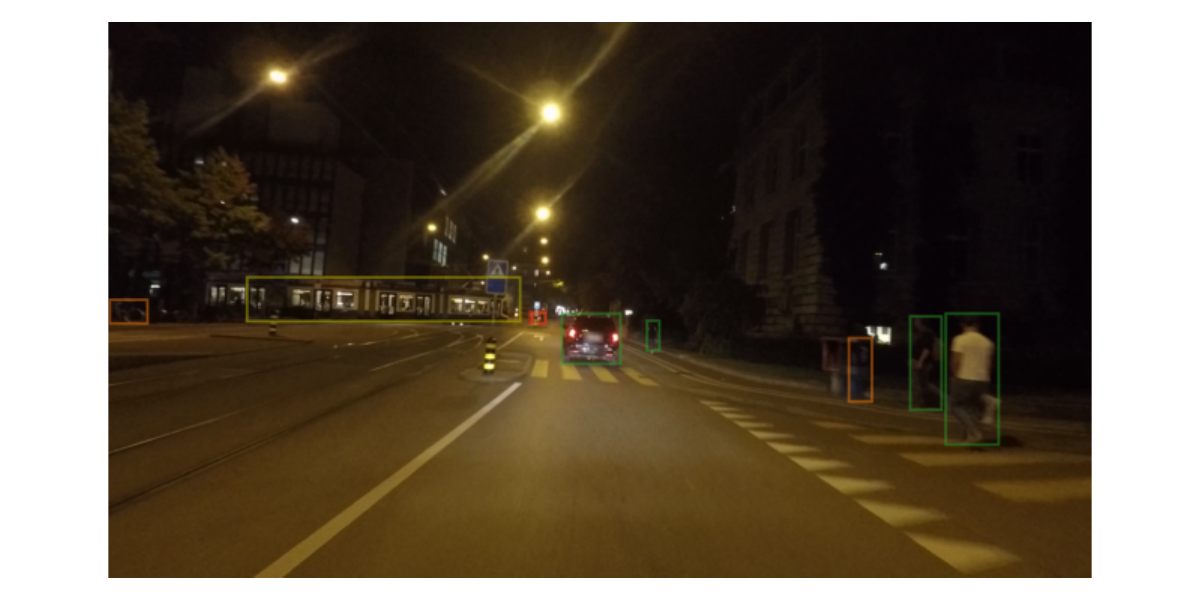}}{}
    \stackunder[0pt]{\includegraphics[width=0.35\textwidth]{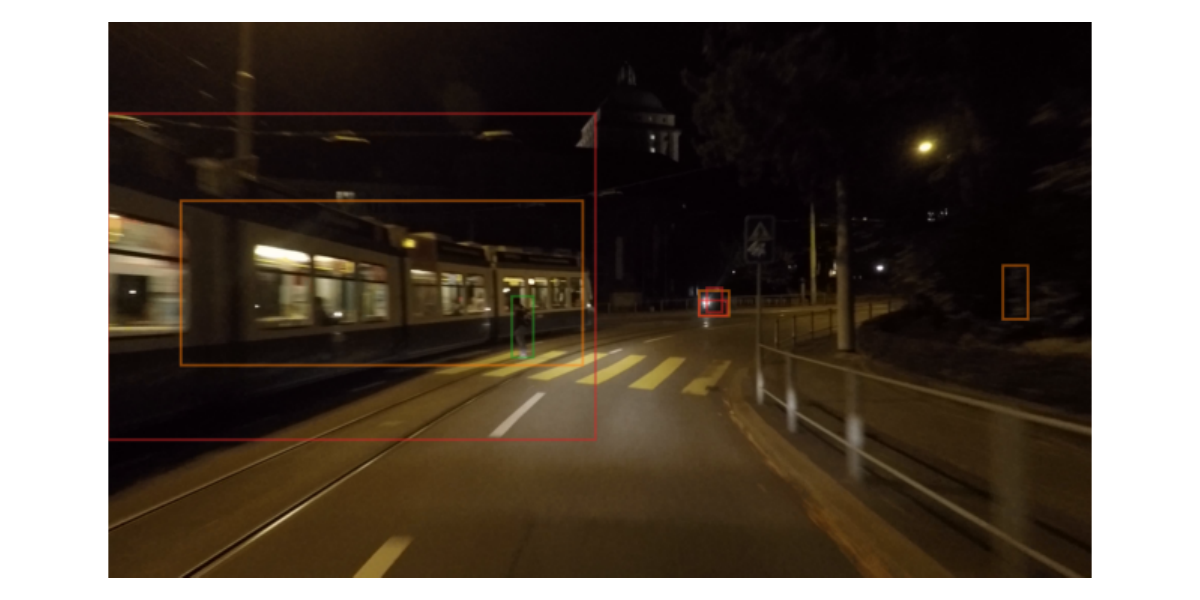}}{Adaptive Teacher}
    \stackunder[0pt]{\includegraphics[width=0.35\textwidth]{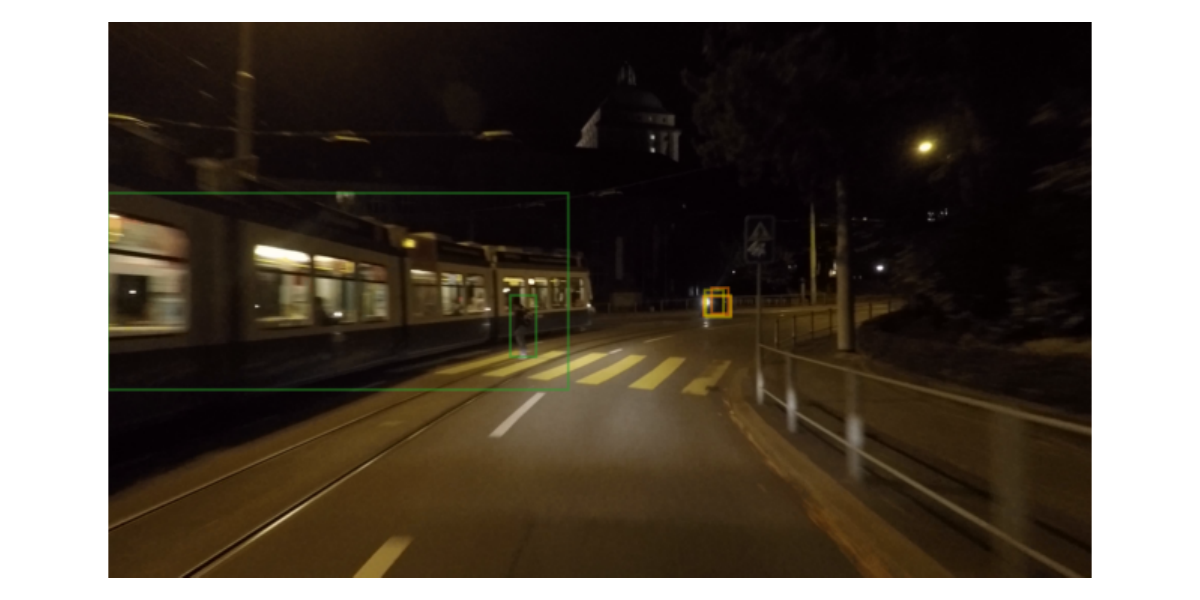}}{DINO Teacher}
\caption{\textbf{Qualitative results on target domain.} We compare Adaptive Teacher (left) to our DINO Teacher (left) on BDD (rows 1 and 2), Foggy Cityscapes (rows 3 and 4) and ACDC Night (rows 5 and 6). {\color{Green}\textbf{Green}}, {\color{Goldenrod}\textbf{Yellow}}, {\color{Orange}\textbf{Orange}} and {\color{Red}\textbf{Red}} indicate true positive, low-confidence positives, false positive, and false negatives respectively. We use a threshold of 0.7 for true positives and false positives.}
\label{fig:visuals}
\end{figure*}

We present qualitative results in \cref{fig:visuals} that compare our approach with the baseline Adaptive Teacher for the transfer to BDD100k, Foggy Cityscapes and ACDC Night. In general, our method performs better for rare classes like trucks and trains and can be better in complex scenarios with overlapping objects. We also generate fewer false positives from wrong classes.

\begin{table*}[t]
\centering
\begin{tabular}{cc|cccccccc|ccc}
\hline \multicolumn{2}{c|}{Run} & Person & Rider & Car & Truck & Bus & Train & Motor & Bicycle & mAP & Avg & Stdev \\
\hline \multirow{3}{*}{Fog} & 1 & 62.6 & 67.2 &	83.0 & 26.1 & 70.3 & 66.3 & 53.3 & 65.4 & 61.8 & \multirow{3}{*}{62.2} & \multirow{3}{*}{0.4} \\
 & 2 & 67.4 & 62.3 & 86.2 & 42.3 & 60.3 & 66.3 & 47.0 & 66.2 & 62.3 \\
 & 3 & 64.9 & 64.1 & 85.4 & 35.2 & 68.6 & 66.3 & 50.7 & 65.6 & 62.6 \\
\hline \multirow{3}{*}{Night} & 1 & 35.6 & 28.0 & 57.5 & 18.5 & -$^{\dagger}$ & 34.0 & 19.5 & 21.8 & 30.7 & \multirow{3}{*}{29.5} & \multirow{3}{*}{1.1} \\
 & 2 & 36.2 & 17.0 & 56.8 & 32.3	& -$^{\dagger}$ & 30.6 & 16.9 & 15.2 & 29.3 \\
 & 3 & 34.9 & 20.3 & 59.5 & 38.7 & -$^{\dagger}$ & 11.6 & 15.2 & 20.2 & 28.6 \\
 \hline \multirow{3}{*}{Rain} & 1 & 45.5 & 4.0 & 76.6 & 58.4 & 37.6 & 21.0 & 50.2 & 4.3 & 37.2 & \multirow{3}{*}{37.0} & \multirow{3}{*}{0.9} \\
 & 2 & 45.6 & 3.8 & 76.0 & 45.1 & 36.5 & 44.9 & 43.6 & 5.9 & 37.7 \\
 & 3 & 40.5 & 19.7 & 76.5 & 37.5 & 37.7 & 14.2 & 61.4 & 0.4 & 36.0 \\
 \hline \multirow{3}{*}{Snow} & 1 & 51.8 & 52.3 & 77.8 & 56.3 & 21.3 & 61.9 & 71.0 & 49.9 & 55.3 & \multirow{3}{*}{55.2} & \multirow{3}{*}{1.0} \\
 & 2 & 53.3 & 64.6 & 78.9 & 56.2 & 20.2 & 61.1 & 70.8 & 44.4 & 56.2 \\
 & 3 & 50.5 & 56.0 & 78.6 & 49.5 & 27.2 & 54.5 & 71.8 & 45.8 & 54.2 \\
\end{tabular}
\caption{\textbf{Full results, Cityscapes to ACDC splits, Adaptive Teacher \cite{li2022cross} baseline.} \textsuperscript{$\dagger$}There are no labels for the Bus class in the night validation split.}
\label{tab:res_acdc_AT_all}
\end{table*}

\begin{table*}[t]
\centering
\begin{tabular}{cc|cccccccc|ccc}
\hline \multicolumn{2}{c|}{Run} & Person & Rider & Car & Truck & Bus & Train & Motor & Bicycle & mAP & Avg & Stdev \\
\hline \multirow{3}{*}{Fog} & 1 & 73.4 & 70.2 & 85.6 & 36.8 & 76.2 & 77.6 & 45.8 & 66.0 & 66.4 & \multirow{3}{*}{68.6} & \multirow{3}{*}{2.2} \\
 & 2 & 66.4 & 75.1 & 85.2 & 40.9 & 89.2 & 80.8 & 56.4 & 72.3 & 70.8 \\
 & 3 & 67.1 & 67.9 & 85.4 & 41.7 & 100.0 & 80.8 & 44.4 & 62.3 & 68.7 \\
\hline \multirow{3}{*}{Night} & 1 & 38.3 & 32.0 & 58.2 & 34.9 & -$^{\dagger}$ & 40.6 & 32.8 & 20.1 & 36.7 & \multirow{3}{*}{36.4} & \multirow{3}{*}{0.4} \\
 & 2 & 41.4 & 27.5 & 58.8 & 37.6 & -$^{\dagger}$ & 50.6 & 23.2 & 16.8 & 36.5 \\
 & 3 & 38.9 & 24.4 & 57.8 & 35.3 & -$^{\dagger}$ & 45.0 & 32.5 & 17.8 & 36.0 \\
 \hline \multirow{3}{*}{Rain} & 1 & 49.3 & 6.4 & 79.9 & 49.8 & 40.0 & 26.2 & 56.3 & 6.5 & 39.3 & \multirow{3}{*}{39.0} & \multirow{3}{*}{0.6} \\
 & 2 & 51.3 & 7.9 & 78.1 & 41.0 & 37.8 & 30.9 & 58.6 & 0.7 & 38.3 \\
 & 3 & 52.3 & 7.7 & 79.2 & 48.5 & 38.1 & 25.9 & 57.6 & 6.0 & 39.4 \\
 \hline \multirow{3}{*}{Snow} & 1 & 50.9 & 70.3 & 78.4 & 55.0 & 20.1 & 62.0 & 60.3 & 60.4 & 57.2 & \multirow{3}{*}{56.9} & \multirow{3}{*}{0.3} \\
 & 2 & 50.5 & 70.3 & 77.4 & 53.0 & 25.2 & 57.2 & 59.0 & 60.6 & 56.7 \\
 & 3 & 55.4 & 64.6 & 76.6 & 55.9 & 20.2 & 66.5 & 63.9 & 50.5 & 56.7 \\
\end{tabular}
\caption{\textbf{Full results, Cityscapes to ACDC splits, proposed DINO Teacher.} \textsuperscript{$\dagger$}There are no labels for the Bus class in the night validation split.}
\label{tab:res_acdc_DT_all}
\end{table*}

{ \vspace{-10pt}
    \small
    \bibliographystyle{ieeenat_fullname}
    \bibliography{main}
}



\end{document}